\documentclass[journal]{IEEEtran}
 
\usepackage{cite}

\usepackage[pdftex]{graphicx}
\usepackage{subfigure}
\usepackage{float}
\usepackage[colorlinks]{hyperref}
\graphicspath{{../pdf/}{../jpeg/}}
\DeclareGraphicsExtensions{.pdf,.jpeg,.png}

\usepackage{tabu}
\usepackage{colortbl}
\usepackage{xcolor}
\usepackage{booktabs}
\usepackage{amsmath}
\usepackage{amsfonts}

\usepackage{fancyhdr}

\begin{document}
	
	\title{Attention Guided Global Enhancement and Local Refinement Network for Semantic Segmentation}
	
	\author{Jiangyun Li,
		Sen Zha, 
		Chen Chen,~\IEEEmembership{Member,~IEEE,}
		Meng Ding,
		Tianxiang Zhang,
		and Hong Yu
		\thanks{J. Li, S. Zha, T. Zhang and H. Yu are with the School of Automation and Electrical Engineering, University of Science and Technology Beijing,
			Beijing 100083, China (e-mail: leejy@ustb.edu.cn, g20198675@xs.ustb.edu.cn, txzhang@ustb.edu.cn, g20198754@xs.ustb.edu.cn)}
		\thanks{C. Chen is with the Center for Research in Computer Vision, University of Central Florida, Orlando, FL 32816 USA (e-mail: chen.chen@crcv.ucf.edu)}
		\thanks{M. Ding is with the Scoop Medical, Houston, TX 77007 USA (e-mail: meng.ding@okstate.edu)
		}
	}
	
	
	\maketitle
	\thispagestyle{fancy}
	\fancyhead{}
	\lhead{}
	\lfoot{\copyright~2022 IEEE.}
	\cfoot{}
	\rfoot{}
	
	\begin{abstract}
		The encoder-decoder architecture is widely used as a lightweight semantic segmentation network. However, it struggles with a limited performance compared to a well-designed Dilated-FCN model for two major problems. First, commonly used upsampling methods in the decoder such as interpolation and deconvolution suffer from a local receptive field, unable to encode global contexts. Second, low-level features may bring noises to the network decoder through skip connections for the inadequacy of semantic concepts in early encoder layers. To tackle these challenges, a Global Enhancement Method is proposed to aggregate global information from high-level feature maps and adaptively distribute them to different decoder layers, alleviating the shortage of global contexts in the upsampling process. Besides, a Local Refinement Module is developed by utilizing the decoder features as the semantic guidance to refine the noisy encoder features before the fusion of these two (the decoder features and the encoder features). Then, the two methods are integrated into a Context Fusion Block, and based on that, a novel Attention guided Global enhancement and Local refinement Network (AGLN) is elaborately designed. Extensive experiments on PASCAL Context, ADE20K, and PASCAL VOC 2012 datasets have demonstrated the effectiveness of the proposed approach. In particular, with a vanilla ResNet-101 backbone, AGLN achieves the state-of-the-art result ($56.23\%$ mean IOU) on the PASCAL Context dataset. The code is available at \url{https://github.com/zhasen1996/AGLN}.
	\end{abstract}
	

	\IEEEpeerreviewmaketitle

	\section{Introduction}
	\IEEEPARstart{S}{emantic} segmentation is a fundamental problem in computer vision, aiming to segment an image into different regions by assigning a predefined class label to each pixel in the image based on their respective semantic categories. It plays a crucial role in various practical applications, such as autonomous driving \cite{cordts2016cityscapes}, computational photography \cite{yoon2015learning}, and human-machine interaction \cite{oberweger2015hands}.
	
	Most of the recent approaches for semantic segmentation are based on Fully Convolutional Network (FCN) \cite{long2015fully}. However, the consecutive downsampling operations, such as pooling and strided convolution, lead to a significant decrease in the spatial details of the original input image. To alleviate this problem, most state-of-the-art methods \cite{chen2014semantic, chen2017deeplab, chen2017rethinking,  zhao2017pyramid, zhang2018context, fu2019dual, zhang2019co, li2019expectation, yuan2020object} discard some of the downsampling operations to maintain relatively high-resolution feature maps and exploit dilated convolutions to enlarge the receptive field, termed Dilated-FCN-based models. Instead of that, another approach carefully designs a decoder on the top of a classification network (i.e. encoder) to gradually reconstruct a high-resolution segmentation map, forming a ``U-shape" Encoder-Decoder architecture \cite{ronneberger2015u, noh2015learning, lin2017refinenet}. Benefiting from the low-resolution intermediate feature maps, the encoder-decoder networks usually require much lower computation and memory costs compared to the widely used Dilated-FCN models. However, there are two major limitations impairing the performance of the encoder-decoder models. 
	
\begin{figure}[!t]
		\centering
		
		\subfigure[Image]{
			\label{Fig.1.sub.1}
			\includegraphics[width=0.118\textwidth]{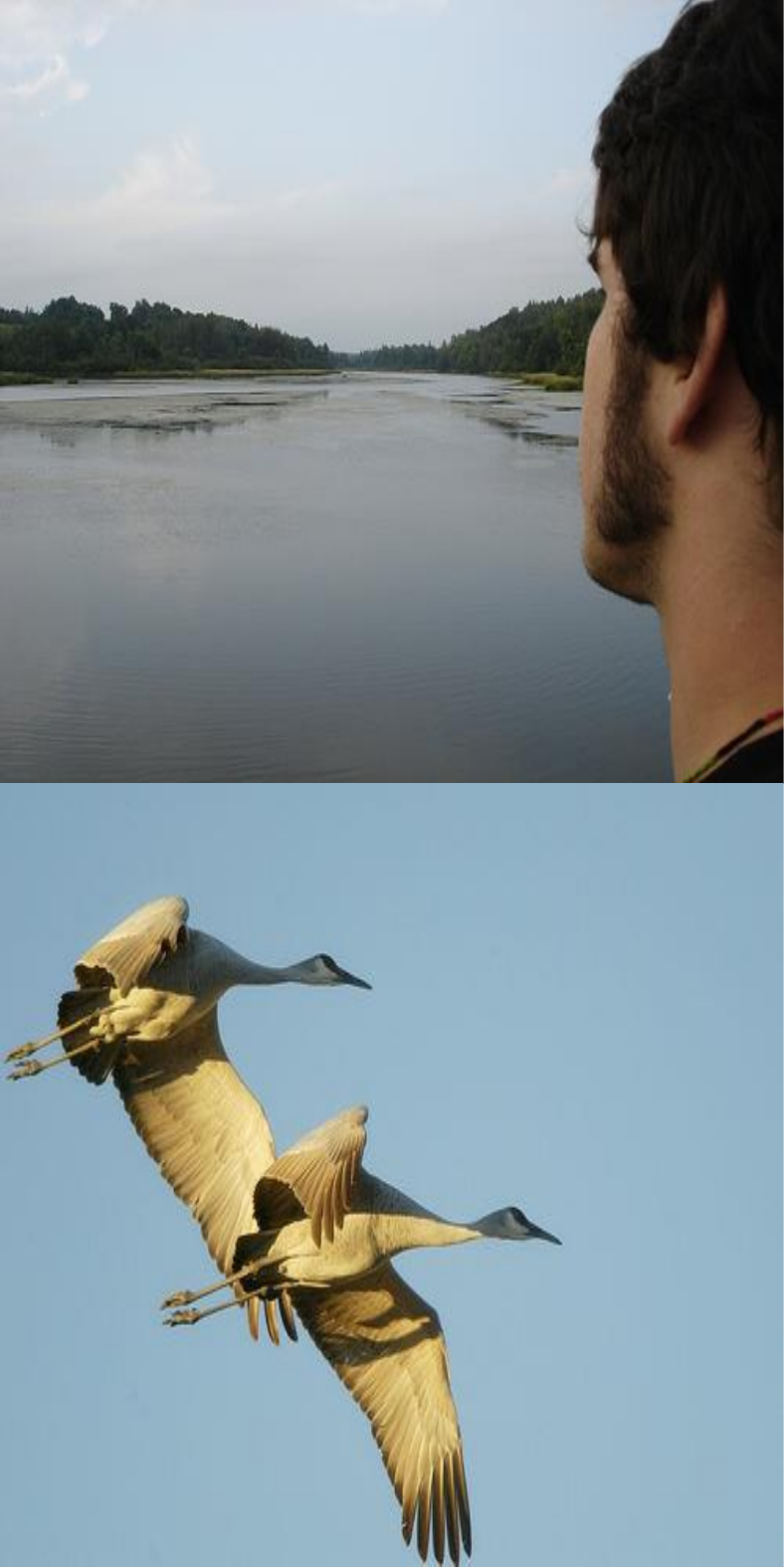}}
		\hspace{-3.2mm}
		\subfigure[GT]{
			\label{Fig.1.sub.2}
			\includegraphics[width=0.118\textwidth]{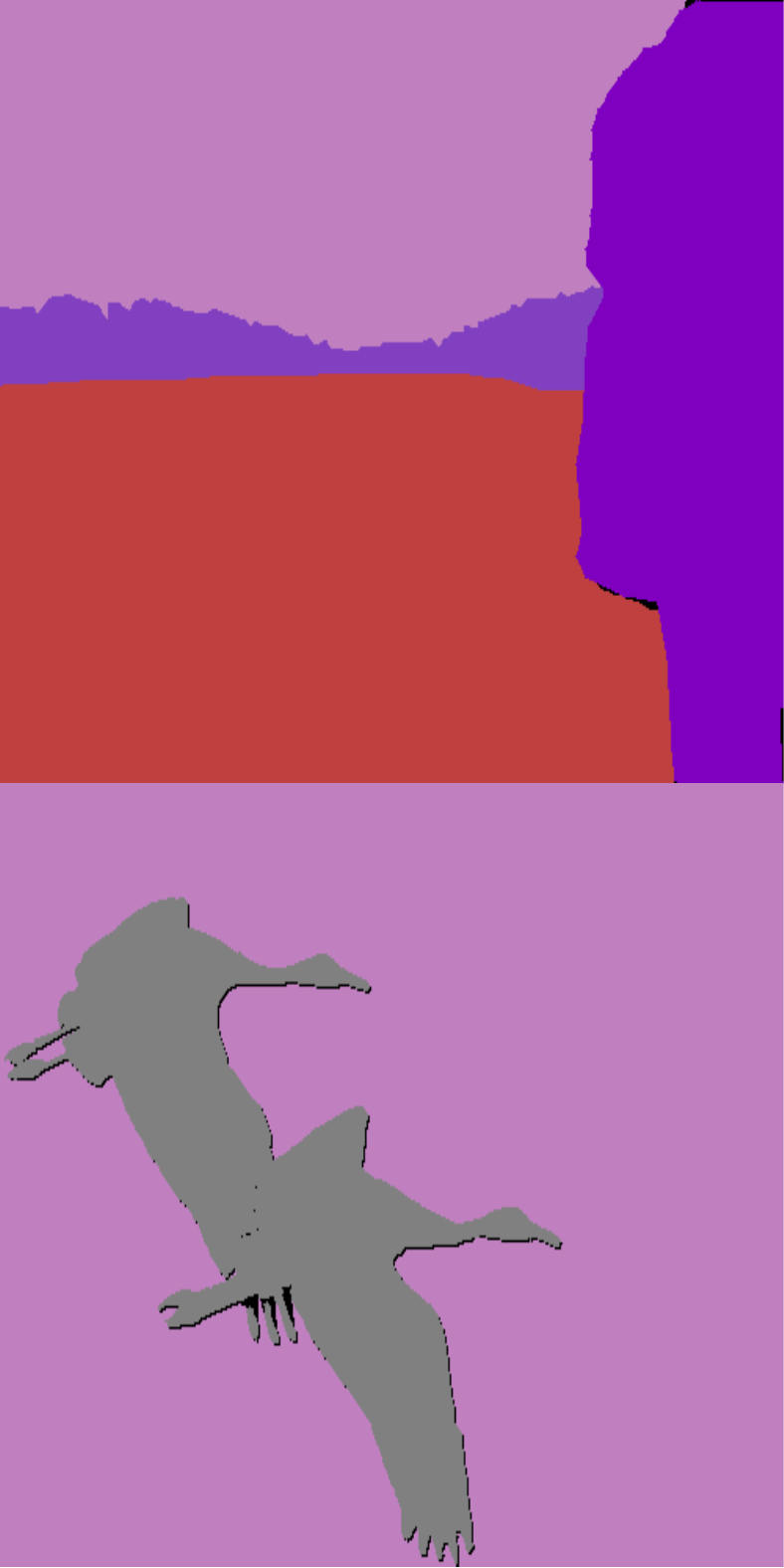}}
		\hspace{-3.2mm}
		\subfigure[Ours]{
			\label{Fig.1.sub.3}
			\includegraphics[width=0.118\textwidth]{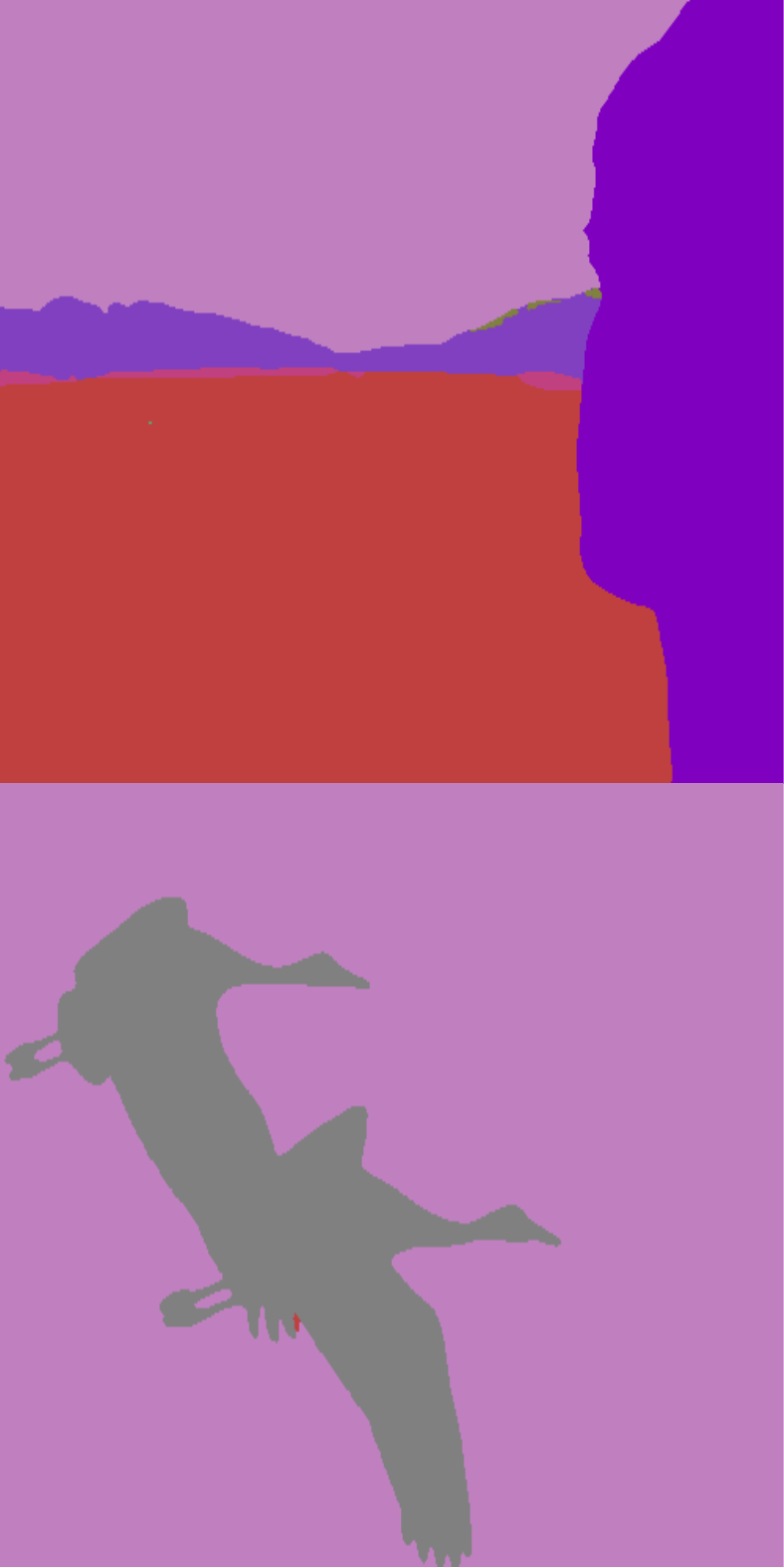}}
		\hspace{-3.2mm}
		\subfigure[FPN]{
			\label{Fig.1.sub.4}
			\includegraphics[width=0.118\textwidth]{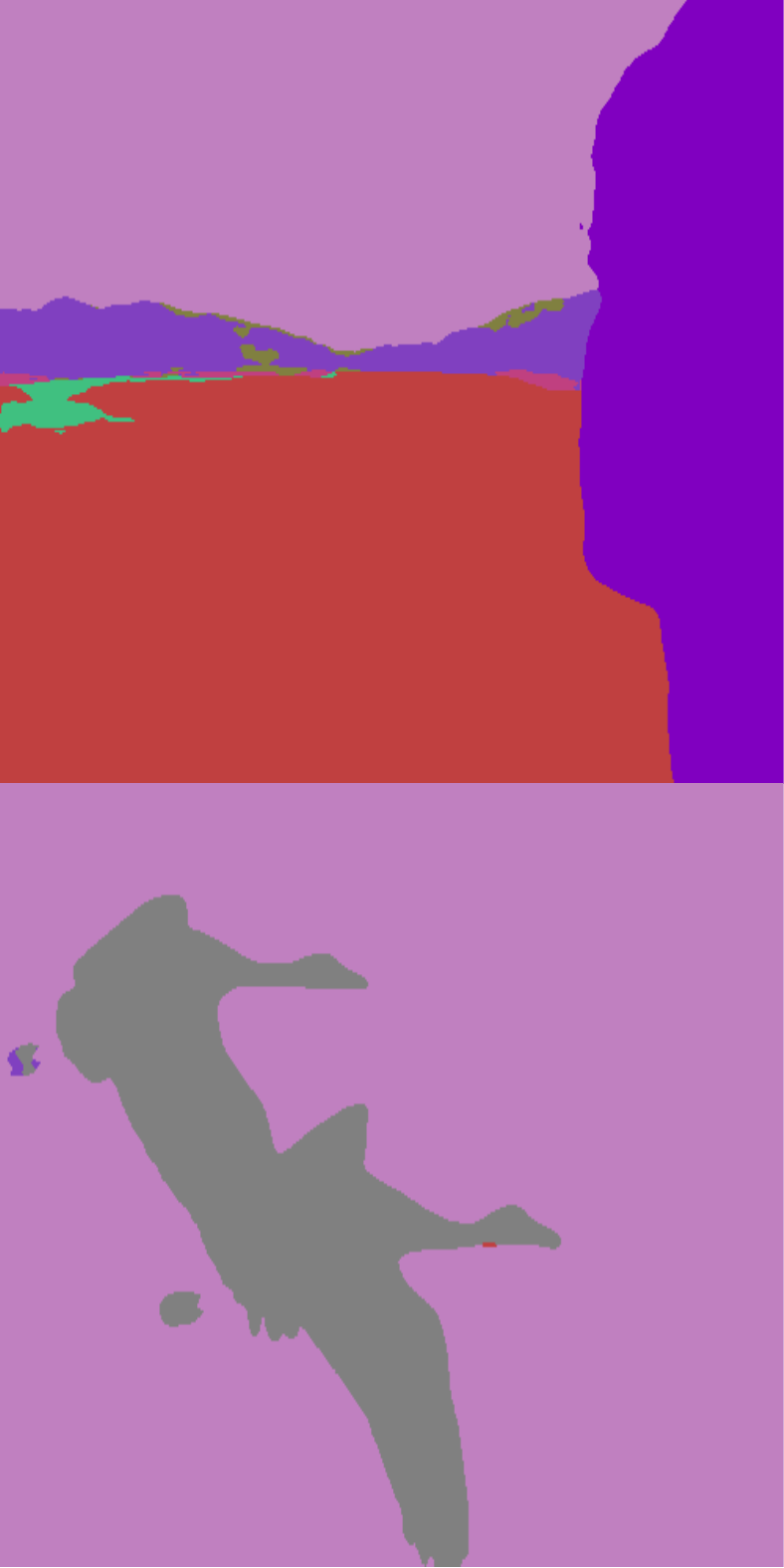}}
		
		\caption{Visualization examples of the segmentation results on PASCAL Context dataset \cite{mottaghi_cvpr14}. From left to right:  (a) Input image, (b) Ground truth, (c) Ours, (d) FPN \cite{lin2017feature}.}
		\label{Fig.1}
		
	\end{figure}

	Firstly, existing decoders mainly utilize bilinear interpolations or deconvolutions to upsample the high-level/low-res feature maps to match the pixel-wise supervision. However, research studies \cite{tian2019decoders, wang2019carafe} revealed that these upsampling methods have limited capability of recovering the pixel-wise prediction accurately.  Specifically, both interpolation and deconvolution only consider the limited neighborhood and are unable to capture long-range contextual information, meaning that the feature representation at each location of the upsampled feature maps is recovered from a limited receptive field. As a result, the absence of global contexts in the upsampling process leads to common misclassification on large-scale objects. As shown in the FPN result (first row of Fig. \ref{Fig.1.sub.4}), an evident mistake (green color) occurs at the top left corner of the water (red color), and this type of error should be avoided. In comparison, our proposed AGLN generates an outstanding result shown in the first row of Fig. \ref{Fig.1.sub.3}.
	
	In addition, low-level features from encoder layers are passed to the decoder through skip connections. However, these encoder feature maps need further refinement since they are unable to provide enough fine-grained details to achieve accurate segmentation boundaries and recognition of small objects \cite{zhang2018exfuse, li2020gated}. Concretely, early encoder convolutions (i.e. sliding windows) traverse all local areas of the image to capture local contexts due to the limited receptive fields. These contexts may be critical for the formulation of semantic concepts in later convolutional layers. However, the massive and confusing local representations (i.e. noises) are not conducive to the recognition of semantic boundaries. Some encoder-decoder networks combine low- and high-level features by simple addition or concatenation operations \cite{ronneberger2015u, milletari2016v, lin2017feature}, drowning critical details in considerable noises, and further resulting in confusion between different objects as well as ignorance of small targets. As shown in the second row of Fig. \ref{Fig.1.sub.4}, FPN cannot recognize the accurate edge of the bird claws since the fine-grained details are lost in the deep network architecture.
	
	The aforementioned two drawbacks make the encoder-decoder models fail to take full advantage of the global semantic information from high-level features and are unable to obtain effective spatial details. Consequently, these problems lead to unacceptable segmentation results, including the misclassification of large-scale objects, rough segmentation boundaries, and ignorance of small targets. To address these challenges, a framework called \textbf{Attention guided Global enhancement and Local refinement Network (AGLN)} is proposed in this paper. Specifically, in terms of the inadequate global information in the feature upsampling process, a \textbf{Global Enhancement Method} is introduced to first selectively aggregate global contexts from high-level features and then adaptively distribute them to the upsampled feature maps. In this way, global category clues are aggregated and delivered to each stage of the decoder, making up for the deficiency of global contexts in the upsampling process and generating the global enhanced decoder feature maps. In addition, utilizing the output of Global Enhancement Method as semantic guidance, a\textbf{ Local Refinement Module} is designed to carefully refine low-level features from the encoder, filtering out the noises and generating more informative local details. After that, these two methods are incorporated into the FPN baseline. The resulting AGLN can provide global enhanced decoder features and local refined spatial details, achieving more effective context fusions in different decoder layers. Consequently, Our AGLN improves the performance of baseline and achieves competitive results on popular benchmarks, including the PASCAL Context \cite{mottaghi_cvpr14}, ADE20K \cite{zhou2017scene}, and PASCAL VOC 2012 \cite{everingham2010pascal} datasets.
	
	Our main contributions can be summarized as follows:
	
	1) We propose a Global Enhancement Method to aggregate global contexts from high-level features and deliver them to different decoder layers, compensating for the absence of global information in the upsampling process.
	
	2) We introduce a Local Refinement Module, providing semantic guidance to refine the encoder features, filtering out the noises before the low- and high-level context fusion.
	
	3) Our proposed AGLN incorporates the aforementioned two methods into the FPN baseline. Extensive experiments show that AGLN achieves state-of-the-art results on PASCAL Context (56.23$\%$ mIOU) and competitive results on ADE20K (45.38$\%$ mIOU) and PASCAL VOC 2012 (84.9$\%$ mIOU).

	\section{Related Work}
	
	In this section, we introduce some popular encoder-decoder models for semantic segmentation and related works on the global context, local context, relation modeling and attention mechanism contributing to the improvements of segmentation performance.
	
	\subsection{Encoder-Decoder Models}
	Based on a U-shape encoder-decoder architecture, different approaches have been explored to improve the segmentation performance. For example, U-Net \cite{ronneberger2015u} and Seg-Net \cite{badrinarayanan2017segnet} employ skip connections on a symmetrical structure to enable the fusion of information from the encoder and decoder. To further exploit the context information along the downsampling process, RefineNet \cite{lin2017refinenet} carefully designs a generic multi-path refinement decoder, fusing coarse high-level semantic features with fine-grained low-level features to generate high-resolution semantic feature maps. In addition, the Feature Pyramid Network (FPN) \cite{lin2017feature} builds a high-level semantic feature pyramid as the decoder to solve the multi-scale object detection problem, guiding the semantic segmentation models at the same time. To summarize, the aforementioned works aim at an effective strategy to utilize different levels of contexts and explore the potential of the encoder-decoder network. For similar purposes, based on the FPN, our AGLN is focused on the enhancement of global contexts and the refinement of local spatial details.
	
	\subsection{Global Context}
	Global context encoding is effective in improving the categorization of large-scale areas and has been well-explored during these years. For instance, PSPNet \cite{zhao2017pyramid} utilizes the global average pooling to collect global contexts and different sub-region representations. EncNet \cite{zhang2018context} introduces a Context Encoding Module to capture the global semantic contexts and predict scaling factors to selectively highlight class-dependent feature maps. ACNet \cite{fu2019adaptive} adopts a global average pooling following a convolution layer to generate the global feature and fuse it with each pixel by an adaptive coefficient measuring the similarity between the global feature and per-pixel feature. OCR \cite{yuan2020object} presents an object-contextual representation approach to aggregate the contexts of all the locations weighted by their degrees belonging to a certain object region. Although these methods improve the performances on different segmentation benchmarks, they completely ignore the inadequacy of global contexts in the upsampling operations. Instead, our Global Enhancement Method is designed to improve the global contexts during the upsampling process.
	
	\subsection{Local Context}
	Local contexts or spatial details are essential to the recognition of small objects and discrimination of segmentation boundaries. Thus, different models have been proposed to exploit the low-level features from the shallow layers. For example, U-Net-based methods \cite{ronneberger2015u, milletari2016v, lin2017feature} often adopt local contexts from low- and middle-level visual features by simple addition or concatenation operations. However, this simple combination of low- and high-level feature maps is less effective because of the gap in semantic levels and spatial resolution between them. Hence, several researches attempt to apply the local contexts in a more effective way. ExFuse \cite{zhang2018exfuse} finds	that introducing semantic information into low-level features and high-resolution details into high-level features is more effective for the later	fusion. RefineNet \cite{lin2017refinenet} utilizes fine-grained features from earlier convolutions to directly refine the deeper layers that capture high-level semantic features through long-range residual connections. ACNet \cite{fu2019adaptive} attaches more local contexts to the locations presenting lower similarities with the global feature and reuses the gated local features multiple times. In addition to the gap between different levels of features, noises from the encoder feature also impair the context fusion. Although some of the previous works \cite{zhang2018exfuse, fu2020scene} point out the problem of noisy low-level features, few of them give a detailed solution to deal with it. In this paper, we propose to refine the noisy encoder features in both spatial and channel dimensions.
	
	\subsection{Relation Modeling and Attention Mechanism}
	The self-attention mechanism has been demonstrated to be capable of modeling relationships between different locations by calculating their similarities with each other. For instance, Non-local Network \cite{wang2018non} and its variants \cite{fu2019dual, fu2020scene, huang2019ccnet, li2019expectation} have demonstrated the significance of the self-attention mechanism in computer vision tasks. Besides, some other models measure feature correlations by learned attention weights. For example, SENet \cite{hu2018squeeze} utilizes the global pooling feature followed by fully connected layers to generate channel weights, selectively highlighting some of the channel maps. CBAM \cite{woo2018cbam} further explores the squeeze-and-excitation module and expands it to the spatial dimension. CPNet \cite{yu2020context} directly supervises the learning of attention weights by developing a Context Prior with the supervision of the Affinity Loss. Similar implementation of the learned attention includes EncNet \cite{zhang2018context}, PSANet \cite{zhao2018psanet}, GANet \cite{zhang2019deep}, and APCNet \cite{he2019adaptive}, etc. Given the demand for global feature extraction and channel relation modeling, the attention mechanism is also well-exploited in our AGLN. Firstly, a learned global attention pooling is adopted in the Global Enhancement Method to adaptively aggregate semantic information in the global spatial space. Additionally, channel attention is employed to model the similarity between the low-level spatial contexts and the enhanced decoder feature, resampling the encoder feature maps in channel dimension.
	
	\section{Proposed Approach} 
	In this section, we first illustrate the major shortcomings of the encoder-decoder segmentation models and then propose two corresponding solutions to overcome these shortages. Specifically, on one hand, a Global Enhancement Method consisting of a Semantic Aggregation Block (SAB) and a Semantic Distribution Module (SDM) is proposed to provide global features inadequate in upsampling process. On the other hand, a Local Refinement Module (LRM) is designed to filter out the noises from the low-level encoder features and generate more informative local details. Based on the proposed methods, a novel AGLN is built to enhance global contexts and refine local textures before the feature fusion in the decoder, leading to improved performance compared to the original encoder-decoder baseline.

	\begin{figure*}[!t]
		\centering
		
		\subfigure[] {
			\label{Fig.2.sub.2}
			\includegraphics[width=0.7\textwidth]{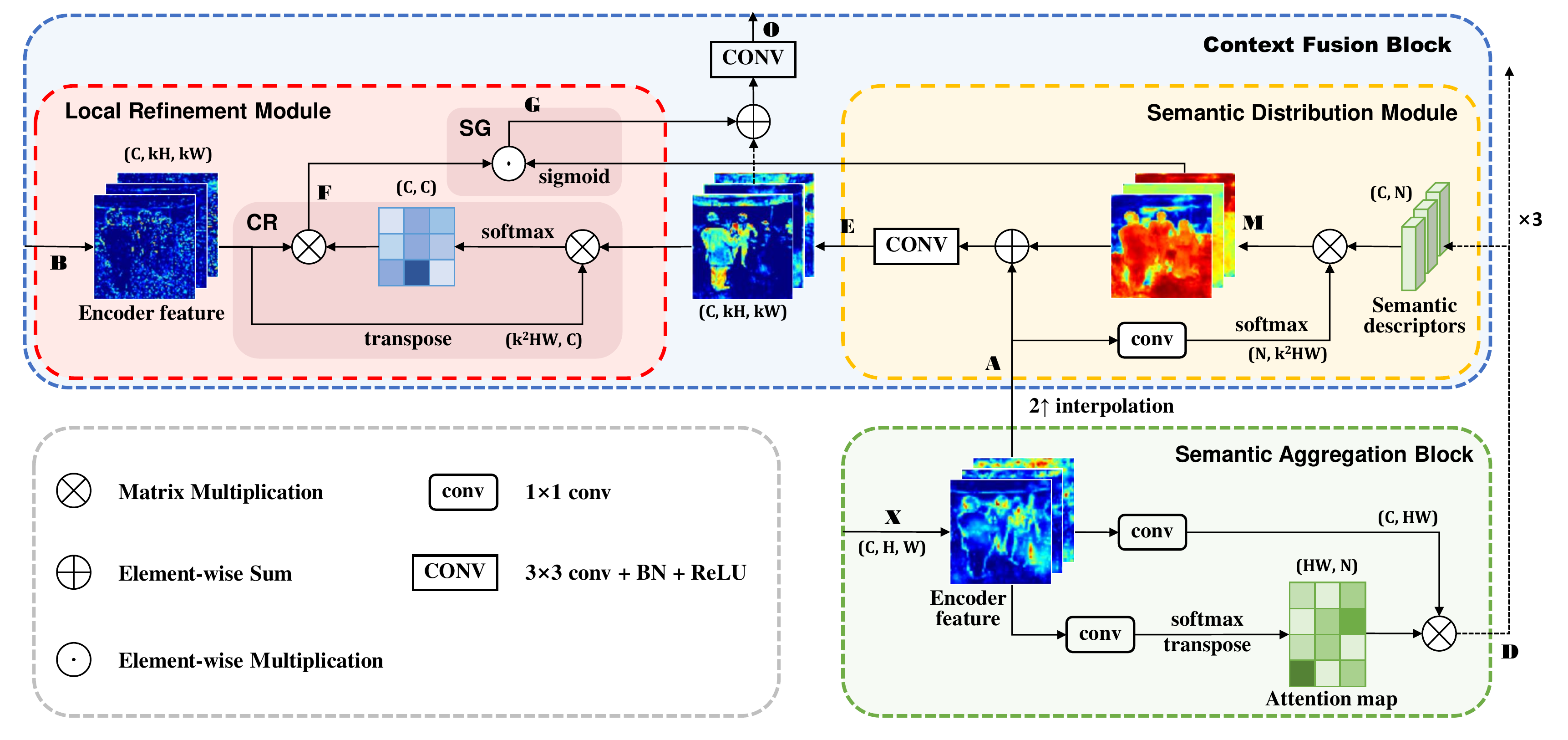}}
		\subfigure[]{
			\label{Fig.2.sub.1}
			\includegraphics[width=0.26\textwidth]{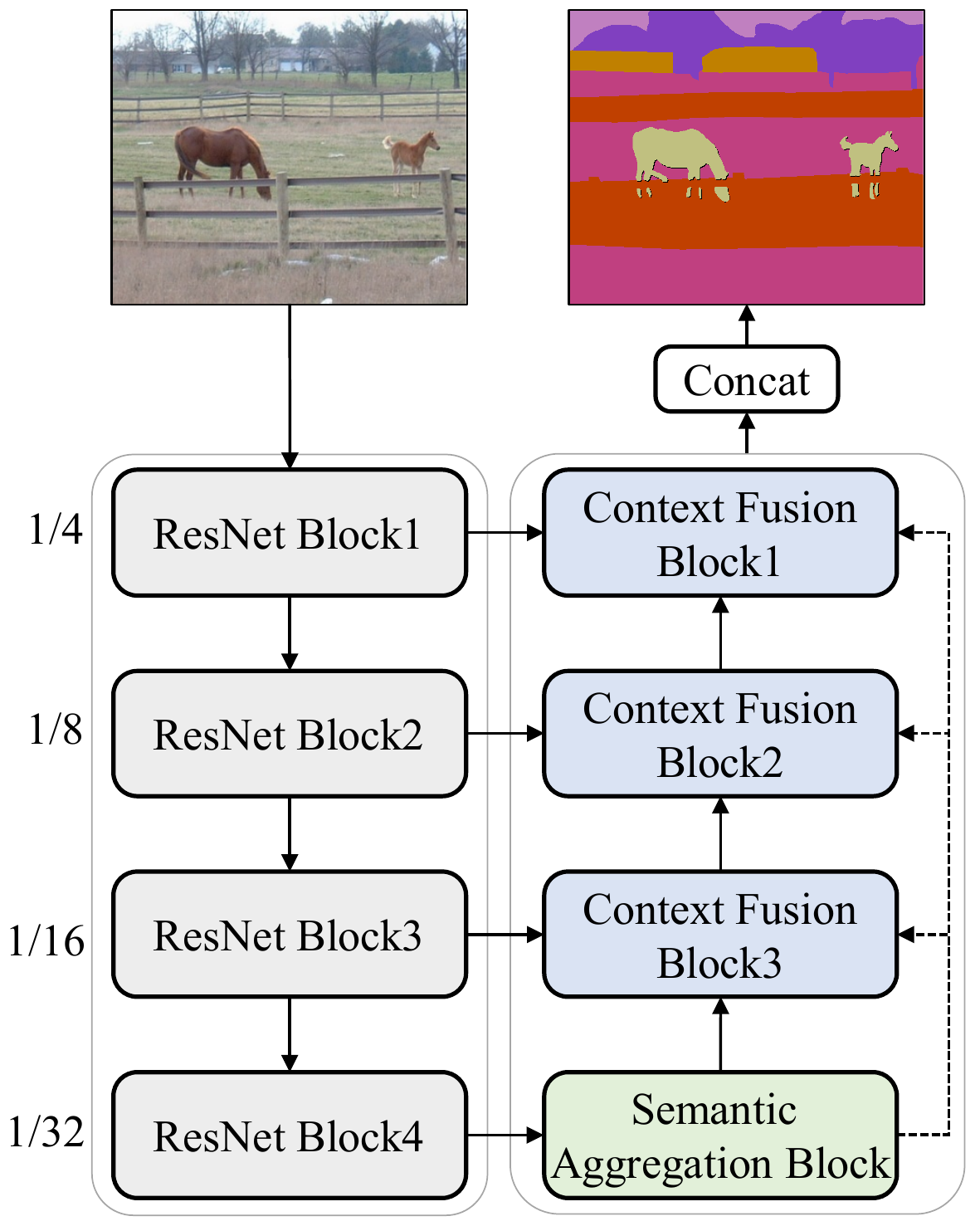}}
		
		\caption{The overall architecture and design details of the proposed AGLN. (a) Details of the proposed Semantic Aggregation Block and Context Fusion Block (consisting of Semantic Distribution Module and Local Refinement Module), (b) Overall architecture of AGLN (best viewed in color). }
		\label{Fig.2}
		
	\end{figure*}

	\subsection{Global Enhancement for Decoder Features}
	
	\subsubsection{Absence of Global Features in Upsampling Process}
	In an encoder-decoder architecture, the deeper layers of encoder contain more high-level semantic information. However, upsampling methods such as nearest interpolation and linear interpolation are incapable of utilizing the global contexts by their hand-crafted kernels in a sub-pixel scope. In addition to the interpolations, another preferred upsampling operator deconvolution also suffers from a limited receptive field (i.e. $3 \times 3$ kernel size). Meanwhile, the deconvolution applies the same kernel across the entire spatial space, despite the various instance-specific contents from different spatial regions.  The aforementioned limitations become the obstacle to the accurate recovery of the pixel-level prediction.
	
	In this work, we aim to break the limits of current upsampling methods and seek an approach to improve the reconstruction of high-resolution segmentation maps. For this purpose, an ideal enhancement method for the upsampling process should be capable of (1) aggregating contextual information in a global space, (2) adaptively updating the representation for each specific location of feature maps, and (3) maintaining computation efficiency. To this end, the Global Enhancement Method is proposed.
	
	\subsubsection{Global Enhancement}
	In our method, the original upsampling operator of the baseline (i.e. bilinear interpolation) is kept. Moreover, we propose to aggregate global features from low-resolution/high-level features and attach them to upsampled decoder feature maps, enabling the restoration of global contexts during the upsampling process.
	
	\textbf{Semantic Aggregation Block}: 
	Firstly, a Semantic Aggregation Block is introduced to \textbf{aggregate global features}. As shown in the green box of Fig. \ref{Fig.2.sub.2}, $ X \in \mathbb{R}^{C \times H \times W} $ denotes the output of the last encoder layer. A \textit{global attention pooling} is applied on $X$ to selectively gather visual primitives from the entire space and generate $N$ semantic descriptors $ D \in \mathbb{R}^{C \times N}$, formulated as:
	\begin{equation}
		D=G_{gather}(X)=X_{feat} \times X_{am}^{\mathsf{T}},
		\label{equation1}
	\end{equation}
	where $G_{gather}(*)$ is the global attention pooling, and ``$\times$'' denotes the matrix multiplication. $X_{feat}=\phi(X;W_{\phi}) \in \mathbb{R}^{C \times HW}$ and $X_{am}=softmax(\theta(X;W_{\theta})) \in \mathbb{R}^{N \times HW}$ are outputs of different $1 \times 1$ convolutional layers ($\phi$ and $\theta$), transforming the input $X$. Specifically, the softmax layer is performed on the spatial dimension ($ HW $) to obtain the attention maps $X_{am}$. 
	
	Due to the diverse representations from different channels ($N$) of the attention map ($X_{am}$), each semantic descriptor is a specific weighted aggregation of global features from the entire spatial space and therefore implicitly encodes a certain semantic region of the input feature map.
	
	\textbf{Semantic Distribution Module}: 
	To \textbf{compensate for the inadequacy of global semantic clues in repeated upsampling operators}, Semantic Distribution Modules are employed to the upsampling process of the encoder-decoder baseline. In the yellow box of Fig. \ref{Fig.2.sub.2}, $A \in \mathbb{R}^{C \times kH \times kW}  (k=2,4,8) $ denotes the decoder feature map gradually upsampled from the lowest-resolution feature map $X$. Given the input feature A, the Semantic Distribution Module first applies a $1\times1$ convolution and a softmax layer in channel dimension, transforming the unique representations from all positions to $kH \times kW$ $N$-dimensional attention vectors. Then, $N$ semantic descriptors are adaptively integrated to each position according to the attention vectors, generating the semantic descriptor map $M \in \mathbb{R}^{C \times kH \times kW} $.
	\begin{equation}
		M=F_{distr}(D,A)=D \times A_{av},
		\label{equation2}
	\end{equation}
	where $F_{distr}(*)$ denotes the semantic distribution function, while $A_{av}=softmax(\varphi(A;W_{\varphi})) \in \mathbb{R}^{N \times kH \times kW} $ are $kH \times kW$ $N$-dimensional attention vectors obtained by channel-wise softmax on the convolved input feature map $A$. 
	
	Next, the semantic descriptor map $M$ is fused with the original decoder feature $A$ by an addition operation followed with convolution, BN, and ReLU layer, generating the output, enhanced decoder feature $E \in \mathbb{R}^{C \times kH \times kW} $. 
	\begin{equation}
		E=\Psi(A+\alpha M;W_{\Psi}).
		\label{equation3}
	\end{equation}
	Note that we multiply $M$ with a learnable scaler parameter $\alpha$  for flexibility, and $\Psi(*;W_{\Psi})$ stands for the convolution together with BN and ReLU layer.
	
	\subsubsection{Discussion}
	Here we provide an in-depth discussion about our Global Enhancement Method.
	
	In the global attention pooling of the Semantic Aggregation Block, the representation of the attention map $X_{am}$ determines the information that a semantic descriptor implicitly encodes. For example, if an attention map has higher activation values on foreground objects (e.g., person, car, and airplane), the semantic descriptor will capture more representations from corresponding foreground objects. By setting the number of semantic descriptors $N$ to a relatively small value (e.g., 64, 128, 256), various global semantic features from the entire spatial space are efficiently aggregated into a set of compact semantic descriptors (The number of semantic descriptors $N$ will be further discussed by an ablation study in the experiment section). After that, Eq. \ref{equation2} of the Semantic Distribution Module formulates the soft attention for feature selection. Hence, the representation of each pixel from the semantic descriptor map $M$ is the adaptive aggregation of the various semantic descriptors. Fusing the semantic descriptor map $M$ with $A$ helps to provide the content-aware global context prior to original decoder features, greatly facilitating the pixel-level classification. Further investigation into the mechanics of the Global Enhancement Method will be elaborated in the experiment Sec. \ref{understanding}.
	
	Furthermore, our Global Enhancement Method is computationally efficient with a complexity of $O(CNHW)$. More specifically, it costs only $CN/2^{18}$ MB memory to store the intermediate result semantic descriptors, meaning saving 128 semantic descriptors for an input data array with 256 channels ($N=128, C=256$, implemented in our method) costs only 0.125 MB memory.
	
	\subsubsection{Comparison} We compare our Global Enhancement Method with other popular global context encoding methods. The self-attention mechanism is applied to capture long-range dependency since its first success in video classification \cite{wang2018non}, and its potential has been well-explored during recent years \cite{fu2019dual, huang2019ccnet, cao2019gcnet}. However, the high computational complexity (quadratic of $HW$) of the self-attention mechanism is still unaffordable in most cases. In comparison, our Global Enhancement Method is more efficient for the linear complexity of $HW$ ($N << HW$). In the process of gathering global information, a great many methods, such as SENet \cite{hu2018squeeze}, CBAM \cite{woo2018cbam}, ACNet \cite{fu2019adaptive} and CGNet \cite{wu2020cgnet}, tend to use a global pooling to generate a single global vector. Our Global Enhancement Method, instead, utilizes a learned global attention pooling (Eq. \ref{equation1}) to get multiple semantic descriptors, thus encoding various semantic representations from different positions in the global feature space. Recently, OCRNet \cite{yuan2020object} also employs the global attention pooling to generate multiple object region representations, however, it needs additional supervision from the ground-truth segmentation to get the object regions (equivalent to the attention map $X_{am}$ in Eq. \ref{equation1}). Moreover, most global context encoding methods design a plugin module and apply it within a single stage of a network. As far as we know, we take the first attempt to enhance the global information during the upsampling process.
	
	\subsection{Local Refinement for Encoder Features}
	\subsubsection{Noisy Encoder Features in Context Fusion}
	In context fusion, the low-level visual primitives encoded by shallow layer features are essential for the accurate prediction of object boundaries and details. However, due to the limited receptive field of the early encoder layers, primary features usually capture massive low-level textures from the entire image, blurring the boundaries between the objects and the background. The needless textures can be regarded as noises in the context fusion since they do few contributions to the recognition of the segmentation boundaries and severely impair the categorization of delicate objects. Compared to the low-level encoder features, representations from the decoder encodes more semantic information, presenting relatively distinct discriminations between different objects and can be even regarded as coarse segmentation maps. Therefore, it is reasonable to utilize the decoder features as the semantic guidance to capture the valuable spatial details and filter out the noises. In the next section, we elaborate on the problem of noisy encoder features from two aspects (i.e. channel dimension and spatial dimension) and propose corresponding solutions.
	
	\subsubsection{Local Refinement}
	A Local Refinement Module is designed to refine the encoder features in both channel (Channel Resampling) and spatial (Spatial Gating) dimensions.
	
	\textbf{Channel Resampling}: Considerable channels of the low-level features present no beneficial information for segmentation and need to be filtered out before the context fusion. Specifically, from the visualization of the encoder features (first two rows of Fig. \ref{Fig.5.sub.2} and \ref{Fig.5.sub.4}), it is observed that a substantial portion of channels (marked by green boxes) shows no boundaries or rough shape of segmentation targets. These channels can be regarded as pure noises for being useless to the recognition of object boundaries. Only a few channels depict clear structure details that truly contribute to the segmentation. To filter out the noisy channels of encoder features, the \textbf{channel attention mechanism is employed to resample the encoder feature maps in channel dimension}, namely Channel Resampling. To be specific, an attention module is built to model the channel dependencies between the noisy encoder features and the semantically rich decoder features. In our design, highly correlated channels of the encoder feature showing stronger dependencies with the decoder feature are emphasized,  and consequently, the representations of specific semantic concepts are improved. Conversely, channels of the encoder feature presenting no discrimination between different objects will be filtered out with a high probability. The implementation details are illustrated in the red box (``CR" denotes the Channel Resampling) of Fig. \ref{Fig.2.sub.2}. Given an encoder feature $B \in \mathbb{R}^{C \times kH \times kW} (k=2,4,8) $ and the enhanced decoder feature $E \in \mathbb{R}^{C \times kH \times kW}$ (the output of Semantic Distribution Module), the channel attention is implemented to get resampled encoder feature map $F$ by
	\begin{equation}
		F=B \times softmax(B \times E^{\mathsf{T}}),
		\label{equation4}
	\end{equation}
	where the softmax operator is applied to obtain the channel attention maps.
	
	\textbf{Spatial Gating}: Low-level features usually encode local details and textures without category awareness. On one hand, the overabundant textural representations outside the target region may blur the object boundaries. On the other hand, detailed features within an object may lead to the miscategorization of other regions. Our insight is to \textbf{utilize the semantic descriptor map $M$ as a spatial gating map to `crop' the valuable details from the region of interest, discarding useless or even harmful textures out the target area.} The semantic descriptor map is chosen for its strong semantic consistency making it more suitable to generate an integrated semantic region compared to the decoder feature map. As illustrated in the red box (``SG" denotes the Spatial Gating) of Fig. \ref{Fig.2.sub.2}, a sigmoid layer is performed on the semantic descriptor map $M$ to obtain the spatial gating map, which is then multiplied with the resampled encoder feature element-wise. The spatial gating process is formulated as
	\begin{equation}
		G=F \cdot sigmoid(M),
		\label{equation5}
	\end{equation}
	where ``$\cdot$'' denotes the element-wise multiplication.
	
	Through these two steps of refinement (Channel Resampling and Spatial Gating), the output of Local Refinement Module, a refined encoder feature $G \in \mathbb{R}^{C \times kH \times kW}$ is finally obtained. 
	
	\subsubsection{Comparison}
	We present comparisons between our Local Refinement Method and some related works. RefineNet \cite{lin2017refinenet} utilizes fine-grained local features to directly refine the deeper layers, while we do the opposite for the insight that deeper layer decoder features can provide semantic guidance to capture more valuable spatial details and filter out the noises. In addition, our design of the Channel Resampling is derived from the Channel Attention Module (CAM) of DANet \cite{fu2019dual}, but the key difference between these two is nonnegligible. Specifically, the CAM performs a self-attention on a single feature map to explicitly model interdependencies between different channels and reweight the channel maps. Our Channel Resampling creatively employs a cross-attention between the encoder and decoder features to enhance fine-grained details and suppress noisy channels from the former (the encoder features).

	\subsection{Context Fusion Block}
	We combine the Semantic Distribution Module with the Local Refinement Module in a Context Fusion Block (blue box of Fig. \ref{Fig.2.sub.2}), enabling it to capture global contexts from semantic descriptors and to refine the encoder features for better local details. Specifically, the refined encoder feature $G$ and the enhanced decoder feature $E$ is combined by element-wise addition, and the result is then fed into a convolution together with BN and ReLU layer, generating the final output of the Context Fusion Block as:
	\begin{equation}
		O=\Phi(E+\beta G;W_{\Phi}),
		\label{equation6}
	\end{equation}
	where $O$ is the final output of the Context Fusion Block. Here the refined encoder feature $G$ is also multiplied with a learnable scalar parameter $\beta$, and $\Phi(*;W_{\Phi})$ denotes the convolution together with BN and ReLU layer.
	
	\subsection{Overall Architecture}
	The overall architecture of the proposed AGLN is shown in Fig. \ref{Fig.2.sub.1}. A pre-trained vallina ResNet \cite{he2016deep} is adopted as the encoder, downsampling the image to 1/32 of the original input size, and then a Semantic Aggregation Block with three Context Fusion Blocks are stacked on top as the decoder.  Following the Feature Pyramid Network \cite{lin2017feature}, the output from different stages of the decoder is upsampled to 1/4 input size and concatenated for final segmentation. In the upsampling process, the Semantic Aggregation Block and three Context Fusion Blocks are employed with different resolutions. Concretely, the Semantic Aggregation Block aggregates global semantic descriptors from the output of the encoder, acting as the first stage of the decoder. After that, the Context Fusion Blocks serve as the subsequent stages, distributing the global semantic descriptors to different stages of decoder features and refining the local details from encoder layers. The output of each decoder block is then interpolated to 1/4 of the original input size for final prediction. Moreover, a pixel-wise cross-entropy loss is applied on the final segmentation result.

	\section{Experiments and Results}
	\label{IV}
	To evaluate our proposed method, comprehensive experiments are conducted on three popular segmentation datasets, including PASCAL Context \cite{mottaghi_cvpr14}, ADE20K \cite{zhou2017scene}, and PASCAL VOC 2012 \cite{everingham2010pascal}. Experimental results demonstrate that the AGLN achieves competitive performances. In the next subsections, we first introduce the datasets, evaluation metrics, and implementation details, and then present our best results compared with state-of-the-art methods on PASCAL Context, ADE20K, and PASCAL VOC 2012 datasets. After that, a series of ablation experiments are performed on the PASCAL Context dataset to validate the effectiveness of our proposed Global Enhancement and Local Refinement. Finally, we propose a lightweight version of the AGLN and conduct experiments to verify its high performance and efficiency.
	
	\subsection{Experimental Settings}
	
	\subsubsection{PASCAL Context}
	The dataset is widely used for scene parsing, which contains 4,998 images for training and 5,105 images for testing. Following \cite{wang2020deep} and \cite{fu2020scene}, our method is evaluated on the most frequent 59 classes (60 classes in total, 59 classes + background). 
	
	\subsubsection{ADE20K}
	The dataset is a scene parsing benchmark containing 20210 images for training, 2000 images for validation, and 3352 images for testing. There are 150 labels in this dataset, including 35 stuff concepts and 115 discrete objects. The unbalanced distribution of labels and diverse scales of images make the dataset more challenging.
	
	\subsubsection{PASCAL VOC 2012}
	The PASCAL VOC 2012 dataset consists of 1,464 images for training, 1,449 for validation, and 1,456 for testing, which is a major benchmark dataset for semantic object segmentation. It includes 20 foreground object classes and one background class. Following the common practice \cite{liu2015semantic, chen2017deeplab, zhang2018context, zhang2019co}, the training set is augmented by additional 10,582 annotated images, namely the PASCAL Aug training set.
	
	\subsubsection{Evaluation Metrics}
	The standard mean Intersection of Union (mIoU) and pixel Accuracy (pixAcc) are chosen as the evaluation metrics. For ablation studies, the single-scale evaluation is performed if not mentioned. While for the best results, following the common practice \cite{zhang2018context, li2020gated, yuan2020object}, the multi-scale evaluation is used to compute mIoU and pixAcc. Specifically, for each input image, we first randomly resize the input image with a scaling factor sampled uniformly from $[0.5, 2.0]$ and then randomly flip the image horizontally. Next, these predictions are averaged to generate the final prediction.

	\subsubsection{Implementation Details}
	The FPN is employed as the baseline, and a pre-trained vanilla ResNet with an output stride of 32 is utilized as the backbone. Note that we apply ResNet-101 as the backbone for the best performance when compared to other state-of-the-art models and ResNet-50 in the ablation study for a shorter experiment time. Following \cite{lin2017feature}, the intermediate outputs of the encoder are passed to $3 \times 3$ convolution with BN and ReLU layers to reduce the number of channels to 256 before feeding into the decoder blocks. Particularly, in this paper, the output of the last encoder layer (i.e. $X$ in Fig. \ref{Fig.2.sub.2}) is fed into Semantic Aggregation Block to obtain the semantic descriptors. After that, three Context Fusion Blocks are stacked on top of the Semantic Aggregation Block, forming the decoder of our AGLN.
	
	We implement our method based on Pytorch. Following \cite{zhao2017pyramid, zhang2018context}, our model is trained with Stochastic Gradient Descent (SGD). During the training phase, a poly learning rate policy is employed where the initial learning rate is multiplied by $ (1-iter/total\_iter)^{0.9} $ after each iteration. The base learning rate is set to 0.001 for PASCAL Context, ADE20K, and 0.0005 for PASCAL VOC 2012. Momentum and weight decay coefficients are set to 0.9 and 0.0001 respectively. The batch size is set to 16 for all datasets. Meanwhile, The training time is set to 240 epochs for PASCAL Context, ADE20K, and 120 epochs for PASCAL VOC 2012. In addition, random scaling, random cropping, and random left-right flipping are applied as data augmentation during training.

	\begin{table}[!t]
		\newcommand{\tabincell}[2]{\begin{tabular}{@{}#1@{}}#2\end{tabular}}
		\renewcommand{\arraystretch}{1.3}
		\caption{Segmentation results on the PASCAL context validation set. D-ResNet denotes the Dilated-ResNet with the output stride = 8 or 16, ResNet denotes the vallina ResNet with the output stride = 32}
		\label{table5}
		\centering
		\begin{tabular}{r |c |c}
			\toprule[1.5pt]
			Method & Backbone & mIoU$\%$ \\
			\hline
			PSPNet\cite{zhao2017pyramid} (CVPR2017) & D-ResNet-101 & 47.8 \\
			EncNet\cite{zhang2018context} (CVPR2018) & D-ResNet-101 & 52.6 \\
			EMANet\cite{li2019expectation} (ICCV2019) & D-ResNet-101 & 53.1 \\
			CGBNet\cite{ding2020semantic} (TIP2020) & ResNet-101 & 53.4 \\
			CPNet\cite{yu2020context} (CVPR2020) & D-ResNet-101 & 53.9 \\
			CFNet\cite{zhang2019co} (CVPR2019) & D-ResNet-101 & 54.0 \\
			HRNetV2\cite{sun2019high} (CVPR2019) & HRNetV2-W48 & 54.0 \\
			ACNet\cite{fu2019adaptive} (ICCV2019) & D-ResNet-101 & 54.1 \\
			GFFNet\cite{li2020gated} (AAAI2020) & D-ResNet-101 & 54.2 \\
			OCRNet\cite{yuan2020object} (ECCV2020) & D-ResNet-101 & 54.8 \\
			DRANet\cite{fu2020scene} (TNNLS2020) & D-ResNet-101 & 55.4 \\
			SETR-MLA\cite{zheng2021rethinking} (CVPR2021) & Transformer & 55.8 \\		
			\tabincell{c}{HRNetV2\cite{wang2020deep} (TPAMI2020)\\ 
				+ OCRNet\cite{yuan2020object} (ECCV2020)} & HRNetV2-W48 & \textbf{56.2} \\
			\hline
			AGLN (Ours) & ResNet-101 & \textbf{56.23} \\
			\bottomrule[1.5pt]
		\end{tabular}
	\end{table}

	\begin{table}[!t]
		\newcommand{\tabincell}[2]{\begin{tabular}{@{}#1@{}}#2\end{tabular}}
		\renewcommand{\arraystretch}{1.3}
		\caption{Segmentation results on the ADE20K Val Set. D-ResNet denotes the Dilated-ResNet with the output stride = 8 or 16, ResNet denotes the vallina ResNet with the output stride = 32}
		\label{table6}
		\centering
		\resizebox{\linewidth}{!}{
			\begin{tabular}{r |c |c | c}
				\toprule[1.5pt]
				Method & Backbone & mIoU$\%$ & PixAcc$\%$\\
				\hline
				PSPNet\cite{zhao2017pyramid} (CVPR2017) & D-ResNet-101 & 43.29 & 81.39\\
				PSANet\cite{zhao2018psanet} (ECCV2018) & D-ResNet-101 & 43.77 & 81.51\\
				EncNet\cite{zhang2018context} (CVPR2018) & D-ResNet-101 & 44.65 & 81.19\\
				CFNet\cite{zhang2019co} (CVPR2019) & D-ResNet-101 & 44.89 & -\\
				CGBNet\cite{ding2020semantic} (TIP2020) & ResNet-101 & 44.90& - \\
				CCNet\cite{huang2019ccnet} (ICCV2019) & D-ResNet-101 & 45.22 & -\\
				OCRNet\cite{yuan2020object} (ECCV2020) & D-ResNet-101 & 45.28 & -\\
				GFFNet\cite{li2020gated} (AAAI2020) & D-ResNet-101 & 45.33 & 82.01\\
				GANet\cite{zhang2019deep} (PR2019) & D-ResNet-101 & 45.36 & \textbf{82.14}\\
				\tabincell{c}{HRNetV2\cite{wang2020deep} (TPAMI2020)\\ 
					+ OCRNet\cite{yuan2020object} (ECCV2020)} & HRNetV2-W48 & \textbf{45.66} & -\\
				\hline
				AGLN (Ours) & ResNet-101 & 45.38 & 81.78\\
				\bottomrule[1.5pt]
			\end{tabular}
		}
	\end{table}

	\begin{table*}[!t]
		\renewcommand{\arraystretch}{1.3}
		\caption{Segmentation results on PASCAL VOC 2012 Test Set. The bold numbers indicate the best one result\\
			 and the gray background highlights the best two results. (without MS COCO dataset pre-trained)}
		\label{table7}
		\centering
		\resizebox{\linewidth}{!}{
			\begin{tabular}{r |c c c c c c c c c c c c c c c c c c c c | c}
				\toprule[1.5pt]
				Method & aero & bike & bird & boat & bottle & bus & car & cat & chair & cow & table & dog & horse & mbike & person & plant & sheep & sofa & train & tv & mIOU$\%$ \\
				\hline
				FCN-8s\cite{long2015fully} & 76.8 & 34.2 & 68.9 & 49.4 & 60.3 & 75.3 & 74.7 & 77.6 & 21.4 & 62.5 & 46.8 & 71.8 & 63.9 & 76.5 & 73.9 & 45.2 & 72.4 & 37.4 & 70.9 & 55.1 & 62.2 \\
				DeepLabv2\cite{chen2017deeplab}  & 84.4 & 54.5 & 81.5 & 63.6 & 65.9 & 85.1 & 79.1 & 83.4 & 30.7 & 74.1 & 59.8 & 79.0 & 76.1 & 83.2 & 80.8 & 59.7 & 82.2 & 50.4 & 73.1 & 63.7 & 71.6 \\
				DeconvNet\cite{noh2015learning} & 89.9 & 39.3 & 79.7 & 63.9 & 68.2 & 87.4 & 81.2 & 86.1 & 28.5 & 77.0 & 62.0 & 79.0 & 80.3 & 83.6 & 80.2 & 58.8 & 83.4 & 54.3 & 80.7 & 65.0 & 72.5 \\
				DPN\cite{liu2015semantic} & 87.7 & 59.4 & 78.4 & 64.9 & 70.3 & 89.3 & 83.5 & 86.1 & 31.7 & 79.9 & 62.6 & 81.9 & 80.0 & 83.5 & 82.3 & 60.5 & 83.2 & 53.4 & 77.9 & 65.0 & 74.1\\
				RefineNet\cite{lin2017refinenet} & 94.9 & 60.2 & 92.8 & \cellcolor{lightgray}\textbf{77.5} & 81.5 & 95.0 & 87.4 & 93.3 & 39.6 & 89.3 & 73.0 & 92.7 & 92.4 & 85.4 & 88.3 & 69.7 & 92.2 & \cellcolor{lightgray}\textbf{65.3} & 84.2 & 78.7 & 82.4\\
				PSPNet\cite{zhao2017pyramid} & 91.8 & 71.9 & 94.7 & 71.2 & 75.8 & 95.2 & 89.9 & 95.9 & 39.3 & 90.7 & 71.7 & 90.5 & 94.5 & 88.8 & 89.6 & 72.8 & 89.6 & \cellcolor{lightgray}64.0 & 85.1 & 76.3 & 82.6 \\
				EncNet\cite{zhang2018context} & 94.1 & 69.2 & \cellcolor{lightgray}\textbf{96.3} & \cellcolor{lightgray}76.7 & \cellcolor{lightgray}\textbf{86.2} & 96.3 & \cellcolor{lightgray}90.7 & 94.2 & \cellcolor{lightgray}38.8 & 90.7 & 73.3 & 90.0 & 92.5 & 88.8 & 87.9 & 68.7 & 92.6 & 59.0 & 86.4 & 73.4 & 82.9 \\
				APCNet\cite{he2019adaptive} & \cellcolor{lightgray}95.8 & 75.8 & 84.5 & 76.0 & 80.6 & \cellcolor{lightgray}96.9 & 90.0 & 96.0 & 42.0 & \cellcolor{lightgray}93.7 & \cellcolor{lightgray}75.4 & 91.6 & 95.0 & \cellcolor{lightgray}90.5 & 89.3 & 75.8 & 92.8 & 61.9 & 88.9 & \cellcolor{lightgray}79.6 & 84.2 \\
				CFNet\cite{zhang2019co} & 95.7 & 71.9 & \cellcolor{lightgray}95.0 & 76.3 & \cellcolor{lightgray}82.8 & 94.8 & 90.0 & 95.9 & 37.1 & 92.6 & 73.0 & \cellcolor{lightgray}93.4 & 94.6 & 89.6 & 88.4 & 74.9 & \cellcolor{lightgray}\textbf{95.2} & 63.2 & \cellcolor{lightgray}\textbf{89.7} & 78.2 & 84.2 \\
				DMNet\cite{he2019dynamic} & \cellcolor{lightgray}\textbf{96.1} & \cellcolor{lightgray}\textbf{77.3} & 94.1 & 72.8 & 78.1 & \cellcolor{lightgray}\textbf{97.1} & \cellcolor{lightgray}\textbf{92.7} & \cellcolor{lightgray}\textbf{96.4} & \cellcolor{lightgray}\textbf{39.8} & 91.4 & \cellcolor{lightgray}\textbf{75.5} & 92.7 & \cellcolor{lightgray}95.8 & \cellcolor{lightgray}\textbf{91.0} & \cellcolor{lightgray}90.3 & \cellcolor{lightgray}76.6 & 94.1 & 62.1 & 85.5 & 77.6 & \cellcolor{lightgray}84.4\\
				\hline
				AGLN (Ours) & 94.6 & \cellcolor{lightgray}77.0 & 93.0 & 71.9 & 81.7 & 95.9 & 90.5 & \cellcolor{lightgray}96.1 & 38.7 & \cellcolor{lightgray}\textbf{95.2} & 74.9 & \cellcolor{lightgray}\textbf{94.2} & \cellcolor{lightgray}\textbf{96.1} & 90.4 & \cellcolor{lightgray}\textbf{90.8} & \cellcolor{lightgray}\textbf{78.6} & \cellcolor{lightgray}94.6 & 59.8 & \cellcolor{lightgray}89.2 & \cellcolor{lightgray}\textbf{83.0} & \cellcolor{lightgray}\textbf{84.9} \\
				\bottomrule[1.5pt]
			\end{tabular}
		}
	\end{table*}

	\subsection{Comparison With State-of-the-Arts}
	\label{4C}
	We compare the best results of our AGLN (based on the vallina ResNet-101 backbone) with recent state-of-the-art models on three popular segmentation benchmarks, PASCAL Context, ADE20K, and PASCAL VOC 2012.
	\subsubsection{Results on PASCAL Context}
	We present the evaluation results of the proposed AGLN on PASCAL Context validation set and compare them with other state-of-the-art methods in Table \ref{table5}. As can be seen, our model achieves the state-of-the-art performance of 56.23$\%$. With the vanilla backbone ResNet-101, our model outperforms Dilated-ResNet-101 based models (e.g., DRAN, GFFNet, and ACNet, etc.). Besides, our result is comparable with HRNetV2 + OCRNet, which is based on a much stronger backbone HRNetV2-W48. The visualization of example results on the PASCAL Context validation set is shown in Fig. \ref{Fig.4} and Fig. \ref{Fig.5}. Compared to FPN, our network generates more accurate segmentation results whether on large-scale background or delicate objects.
	
	\subsubsection{Results on ADE20K}
	Experiments are also conducted on the ADE20K dataset to validate the performance of our AGLN. Quantitative results (evaluated by both PixelAcc and mIoU) are shown in Table \ref{table6}. With vallina ResNet-101, our AGLN achieves a competitive performance of 45.38$\%$ mIoU and 81.78$\%$ PixAcc compared to other state-of-the-art models. Note that the HRNetV2 + OCRNet outperforms our network with a much stronger backbone HRNetV2-W48. Without the HRNetV2 backbone, OCRNet shows an inferior performance when compared with our method. The visualization of the example results from FPN and our AGLN is presented in Fig. \ref{Fig.6}. The main contrasts marked by the white rectangles show that our results present stronger classification consistency within an object and more accurate boundaries.
	
	\subsubsection{Results on PASCAL VOC 2012}
	The performance of the proposed AGLN is also evaluated on the PASCAL VOC 2012 dataset. Following \cite{zhang2018context, zhang2019co}, our model is first trained on the PASCAL Aug training data for 120 epochs and then fine-tuned on the original PASCAL VOC 2012 dataset for another 120 epochs. As shown in Table \ref{table7}, AGLN achieves 84.9$\%$ mIOU without COCO pre-training and performs best in many categories, from large objects (e.g., train, cow, horse, and tv) to small, delicate ones (e.g., bike, cat, dog, and plant).

	\begin{table}[!t]
		\renewcommand{\arraystretch}{1.3}
		\caption{Ablation study for the Global Enhancement Method. GEM represents the Global Enhancement Method while GAP-GEM replaces the global attention pooling to global average pooling in GEM. N denotes the number of semantic descriptors}
		\label{table1}
		\centering
		\begin{tabular}{c c c |c}
			\toprule[1.5pt]
			Method & Backbone & N & mIoU$\%$ \\
			\hline
			FPN (w/o GEM) & ResNet-50 & & 46.91 \\
			\hline
			AGLN$^{-}$(w/ GAP-GEM) & ResNet-50 &  & 48.17 \\
			AGLN$^{-}$(w/ GEM) & ResNet-50 & 1 &  47.93\\
			AGLN$^{-}$(w/ GEM) & ResNet-50 & 64 &  50.98\\
			AGLN$^{-}$(w/ GEM) & ResNet-50 & 128 & \textbf{51.24}\\
			AGLN$^{-}$(w/ GEM) & ResNet-50 & 256 & 50.88 \\
			\bottomrule[1.5pt]
		\end{tabular}
	\end{table}
	
	\begin{table}[!t]
		\renewcommand{\arraystretch}{1.3}
		\caption{Ablation study for Local Refinement Module. CR represents Channel Resampling and SG represents Spatial Gating}
		\label{table2}
		\centering
		\begin{tabular}{c c c c |c}
			\toprule[1.5pt]
			Method & Backbone & CR & SG & mIoU$\%$ \\
			\hline
			FPN & ResNet-50 &  & & 46.91 \\
			\hline
			AGLN$^{-}$ & ResNet-50 &  &  &  51.24 \\
			AGLN$^{-}$ & ResNet-50 & $\surd$ &  &  51.50 \\
			AGLN$^{-}$ & ResNet-50 &  & $\surd$ & 51.58 \\
			AGLN & ResNet-50 & $\surd$ & $\surd$ & \textbf{ 51.98} \\ 		
			\bottomrule[1.5pt]
		\end{tabular}
	\end{table}

	\begin{table}[!t]
		\renewcommand{\arraystretch}{1.3}
		\caption{Ablation study for learnable parameters $\alpha$ and $\beta$. 1 means the parameter is fixed to a constant value of 1, $\ell$ means the parameter is learnable}
		\label{table3}
		\centering
		\begin{tabular}{ c c c c |c}
			\toprule[1.5pt]
			Method & Backbone & $\alpha$ & $\beta$ & mIoU$\%$ \\
			\hline
			AGLN & ResNet-50 & $\ell$ & $\ell$ & \textbf{51.98} \\
			AGLN & ResNet-50 & 1 & $\ell$ & 51.20 (-0.78) \\
			AGLN & ResNet-50 & $\ell$ & 1 & 51.68 (-0.30) \\
			AGLN & ResNet-50 & 1 & 1 & 50.93 (-1.05) \\
			\bottomrule[1.5pt]
		\end{tabular}
	\end{table}

	\begin{figure*}[!t]
		\centering
		\subfigure[]{
			\label{Fig.4.sub.1}
			\includegraphics[width=0.14\textwidth]{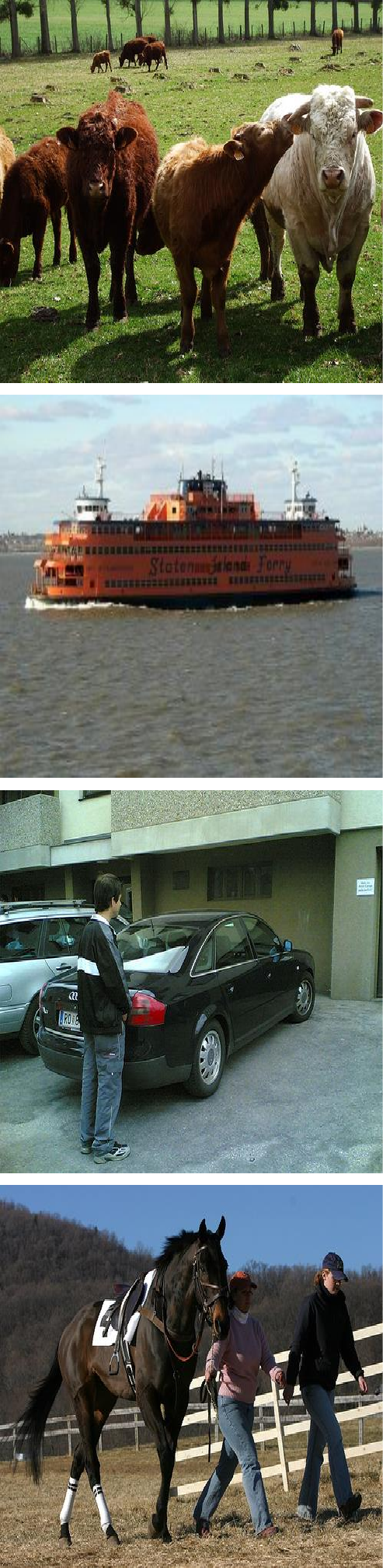}}
		\hspace{-3mm}
		\subfigure[]{
			\label{Fig.4.sub.3}
			\includegraphics[width=0.14\textwidth]{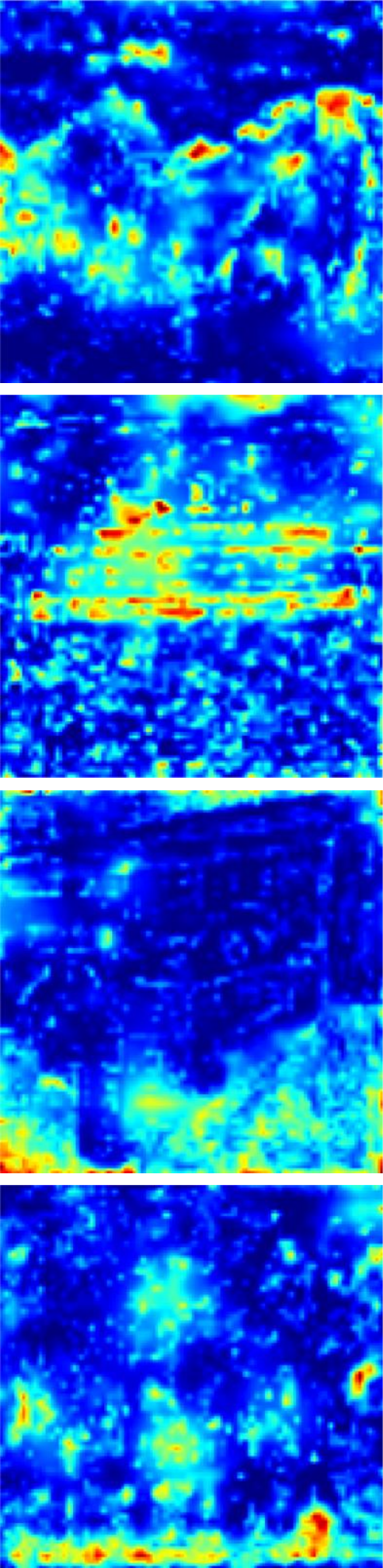}}
		\hspace{-3mm}
		\subfigure[]{
			\label{Fig.4.sub.2}
			\includegraphics[width=0.14\textwidth]{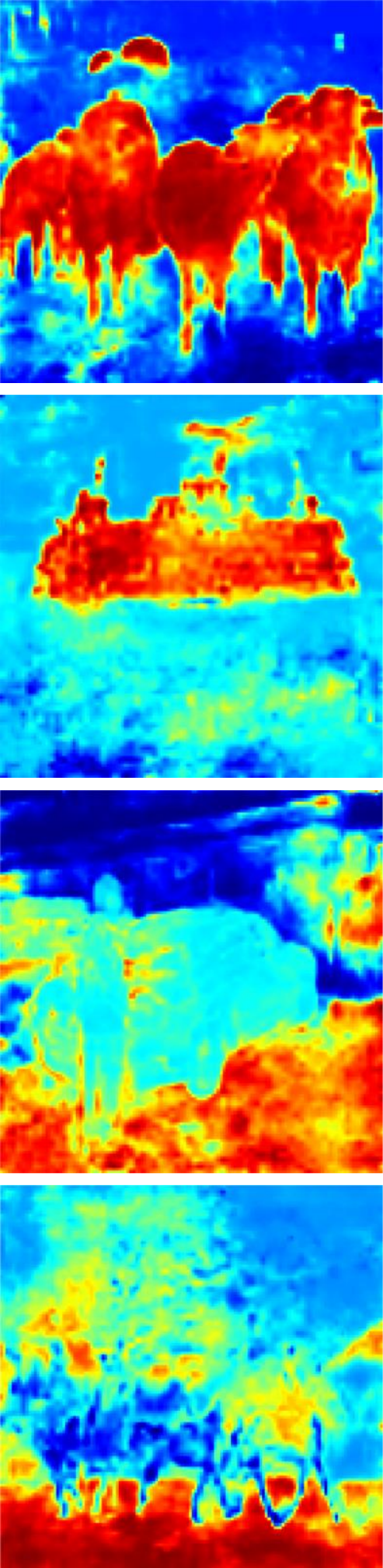}}
		\hspace{-3mm}
		\subfigure[]{
			\label{Fig.4.sub.4}
			\includegraphics[width=0.14\textwidth]{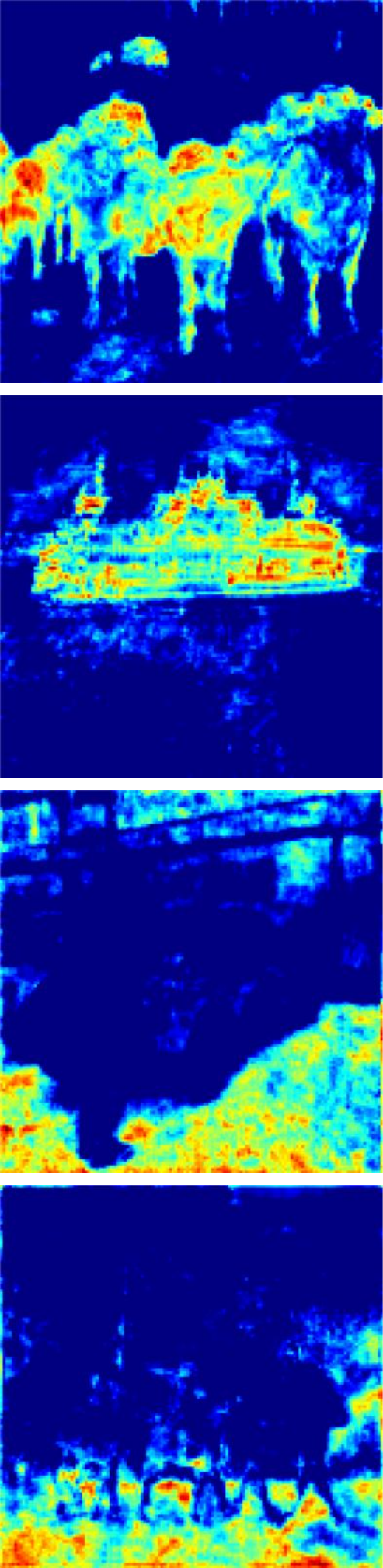}}
		\hspace{-3mm}
		\subfigure[]{
			\label{Fig.4.sub.5}
			\includegraphics[width=0.14\textwidth]{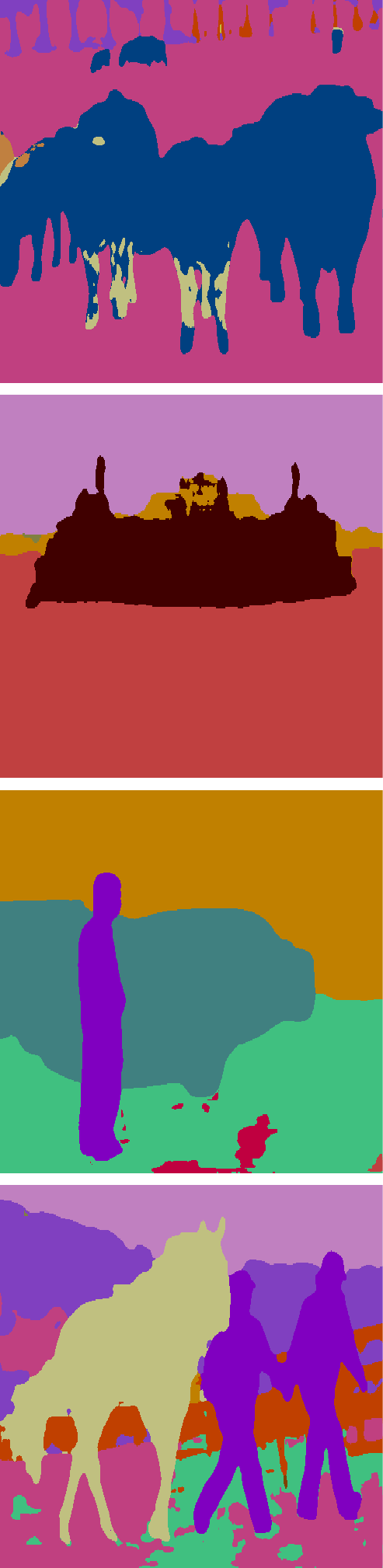}}
		\hspace{-3mm}
		\subfigure[]{
			\label{Fig.4.sub.6}
			\includegraphics[width=0.14\textwidth]{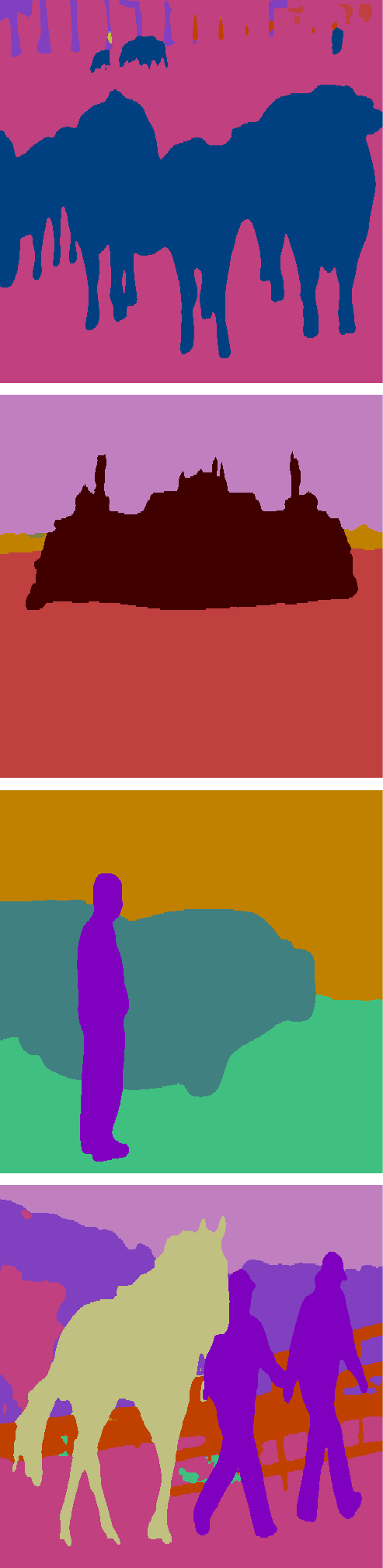}}
		\hspace{-3mm}  
		\subfigure[]{
			\label{Fig.4.sub.7}
			\includegraphics[width=0.14\textwidth]{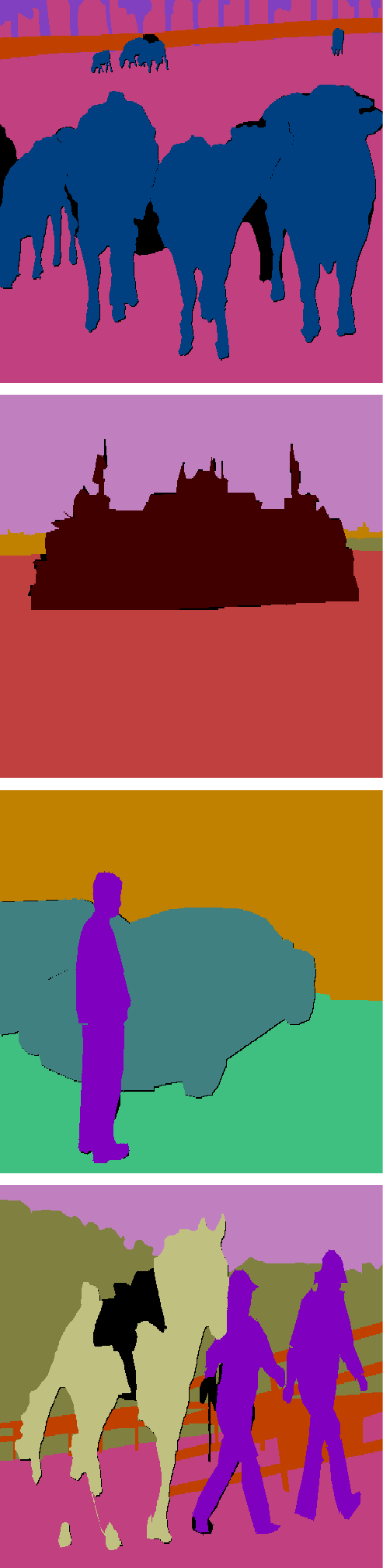}}
		
		\caption{The visualization of feature maps and segmentation results from FPN and our AGLN on the PASCAL Context dataset. From left to right: (a) Input images; (b) Decoder features; (c) Semantic descriptor maps; (d) Enhanced decoder features (the fusion results of decoder features and semantic descriptor maps); (e) FPN results; (f) Our results; (g) Ground truth. (The feature maps are from the last stage of the decoder)}
		
		\label{Fig.4}
	\end{figure*}

	\subsection{Ablation Studies}
	\label{4D}
	Ablation studies are performed on the PASCAL Context validation set to verify the effectiveness of different components of our network. All experiments are conducted on the ResNet-50 backbone. 
	
	\subsubsection{Ablation Study for Global Enhancement Method}
	\label{4D1}
	To evaluate the effectiveness of our proposed Global Enhancement Method (consisting of Semantic Aggregation Block and Semantic Distribution Modules), the performance of the FPN with and without Global Enhancement is compared. To this end, we apply the Global Enhancement Method on the FPN baseline. The resulting AGLN$^{-}$ shares a similar structure with AGLN while its decoder is composed of a Semantic Aggregation Block and three Semantic Distribution Modules. As shown in Table \ref{table1}, the Global Enhancement Method improves the performance significantly. Compared with the FPN baseline, our  AGLN$^{-}$ obtains a result of more than 51$\%$ in mIoU, bringing at least 4$\%$ improvement. In our design, the Semantic Aggregation Block plays a vital role in capturing various global semantic features, the performance witnesses an obvious decline if we replace the global attention pooling in SAB with a global average pooling (GAP-GEM in Table \ref{table1}).

	\subsubsection{Ablation Study for Number of Semantic Descriptors}
	\label{4D2}
	Different numbers of semantic descriptors ($N$) are applied in experiments for a comprehensive comparison to show their effect on the segmentation results. As shown in Table \ref{table1}, compared to 64, increasing the number of semantic descriptors to 128 could improve the performance (i.e. the bold number 51.24$\%$). However, continue raising it to 256 gains no further improvement for the segmentation results. Considering the limited segmentation classes in an image, a compact set of semantic descriptors may lead to an easier categorization. In addition, if $N$ is set to 1, only one semantic descriptor is obtained and the attention pooling is degraded to a global average pooling in performance. Our later experiments are based on the best setting of 128 semantic descriptors ($N$ = 128).

	\subsubsection{Ablation Study for Local Refinement Module}
	Based on AGLN$^{-}$, further experiments are conducted to evaluate the effectiveness of the Local Refinement Module. As shown in Table \ref{table2}, the Channel Resampling (CR) and Spatial Gating (SG) separately improve the performance to 51.50$\%$ and 51.58$\%$. When we combine these two methods into a Local Refinement Module and form the complete AGLN, the result further goes up to 51.98$\%$. The comparison of segmentation results between our AGLN and the baseline in Fig. \ref{Fig.5} shows the effectiveness of the Local Refinement Module. Apparently, Our AGLN generates more accurate segmentation edges compared to the FPN.

	\subsubsection{Ablation Study for Learnable Parameters $\alpha$ and $\beta$}
	When fusing semantic descriptor maps and refined encoder features into the decoder, we respectively multiply them with learnable parameters $\alpha$ and $\beta$ for flexibility. Experiments are also performed to investigate their effect on the segmentation results. In practice, the initial value of $\alpha$ is set to 0 to avoid the disturbance from semantic descriptor maps at the beginning of training, and $\beta$ is initialized as 1. As shown in Table \ref{table3},  without the learnable $\alpha$ and $\beta$, the performance of AGLN decreases by different margins (-0.78\% and -0.30\% respectively). When $\alpha$ and $\beta$ are both fixed to 1 during the training process, the mIoU of AGLN drops down by 1\%.

	\subsubsection{Densely Connected Decoder and Multi-scale Evaluation}
\begin{figure}[!t]
		\centering
		
		\subfigure[]{
			\label{Fig.3.sub.1}
			\includegraphics[width=0.162\textwidth]{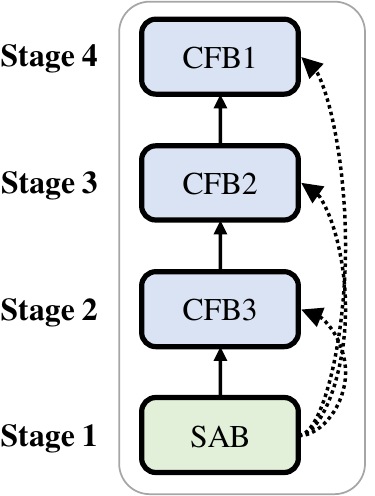}}
		\hspace{-3mm}
		\subfigure[] {
			\label{Fig.3.sub.2}
			\includegraphics[width=0.3\textwidth]{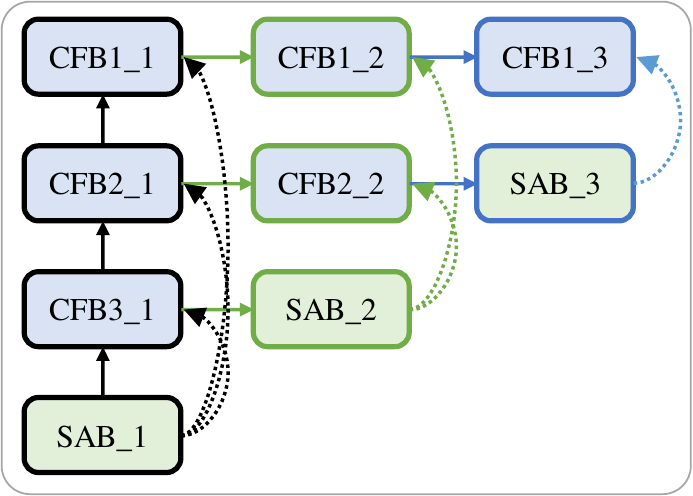}}
		
		\caption{Different decoder implementations of our network: (a) Straight connection, (b) Dense connection of SAB and CFBs.}
		\label{Fig.3}
		
	\end{figure}

	Inspired by the dense connection in \cite{huang2017densely} and feature reuse in \cite{fu2019adaptive}, a further improvement of our network is developed by applying the Semantic Aggregation Block and Context Fusion Blocks multi-times in a dense connection way. As shown in Fig. \ref{Fig.3}, compared to the straight connection, the densely connected implementation has two advantages: (1) global semantic information from the second and third stages of decoder features could be well-exploited through the repeatedly applied Semantic Aggregation Blocks (i.e. ``SAB\_2" and ``SAB\_3" in Fig. \ref{Fig.3.sub.2}); (2) the spatial details from encoder features are further enhanced in such a recurrent learning process through the consecutive Context Fusion Blocks (i.e. ``CFB1\_2", ``CFB1\_3" and ``CFB2\_2" in Fig. \ref{Fig.3.sub.2}). As shown in Table \ref{table4}, the implementation of the densely connected decoder improves the performance of AGLN to 52.45$\%$, which is a significant improvement over the baseline by 5.54$\%$. Moreover, following \cite{zhang2018context, fu2019dual}, multi-scale testing is performed on AGLN to further improve the mIoU to 53.42$\%$. Our best results in Sec. \ref{4C} are obtained by the best implementation described in this section.

	\begin{figure*}[!t]
		
		\centering
		\subfigure[]{
			\label{Fig.5.sub.1}
			\includegraphics[width=0.426\textwidth]{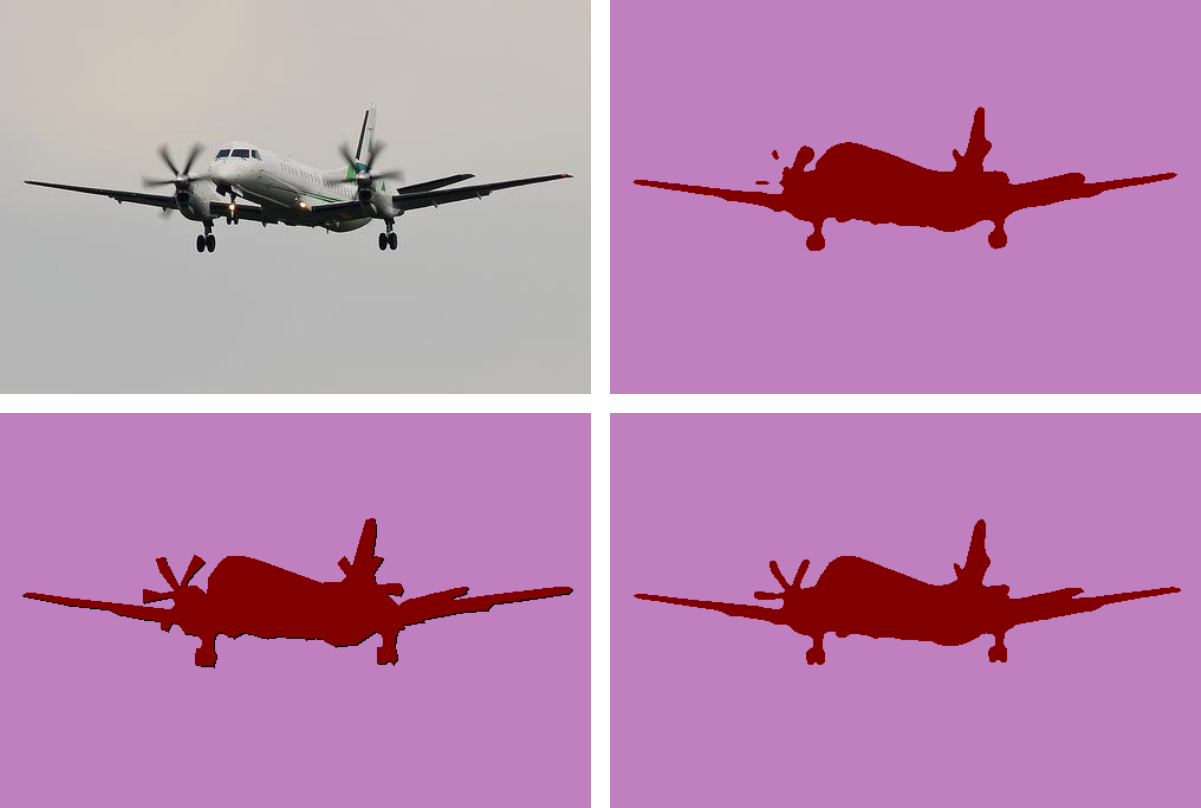}}
		\hspace{-3mm}
		\subfigure[] {
			\label{Fig.5.sub.2}
			\includegraphics[width=0.559\textwidth]{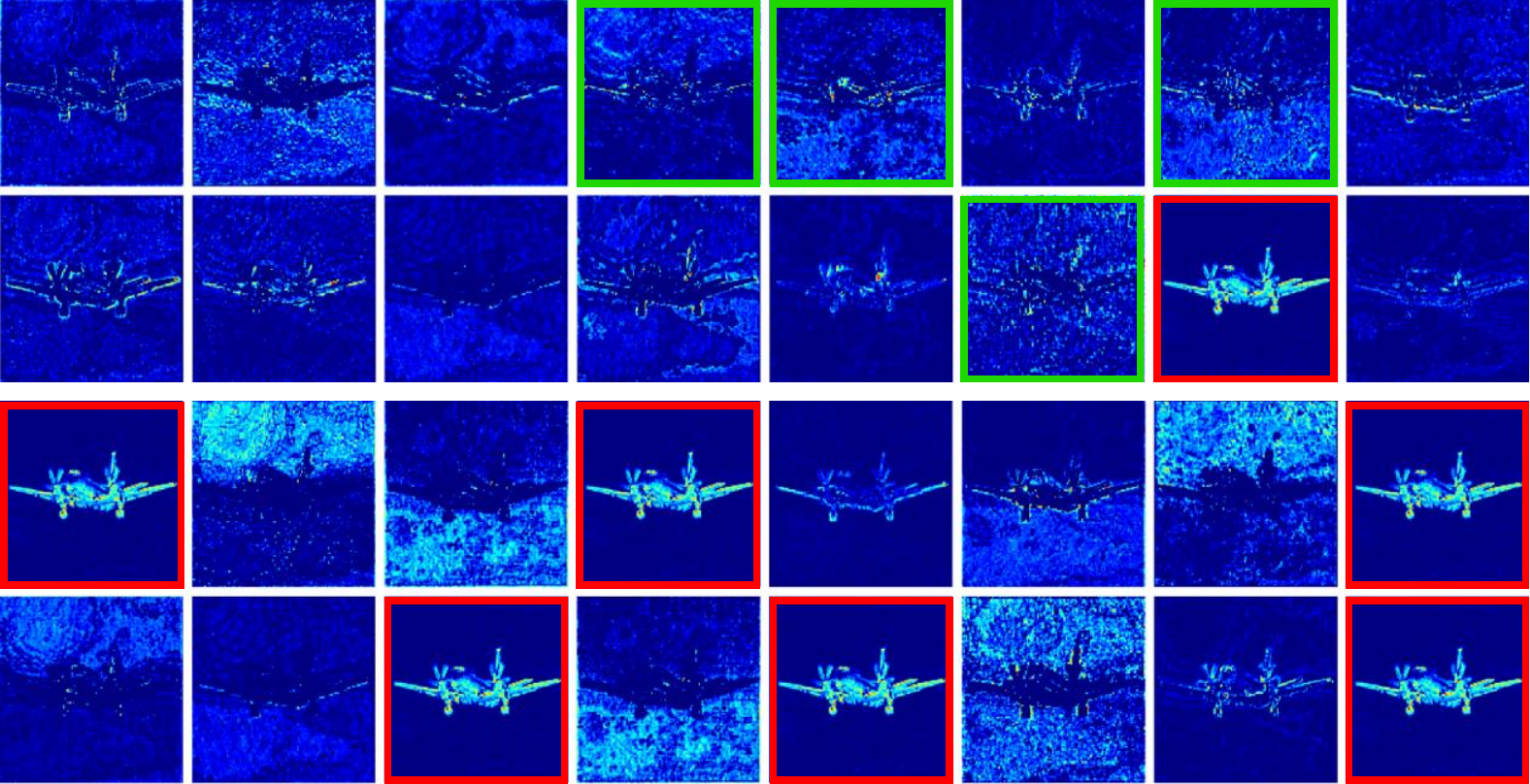}}
		\subfigure[]{
			\label{Fig.5.sub.3}
			\includegraphics[width=0.426\textwidth]{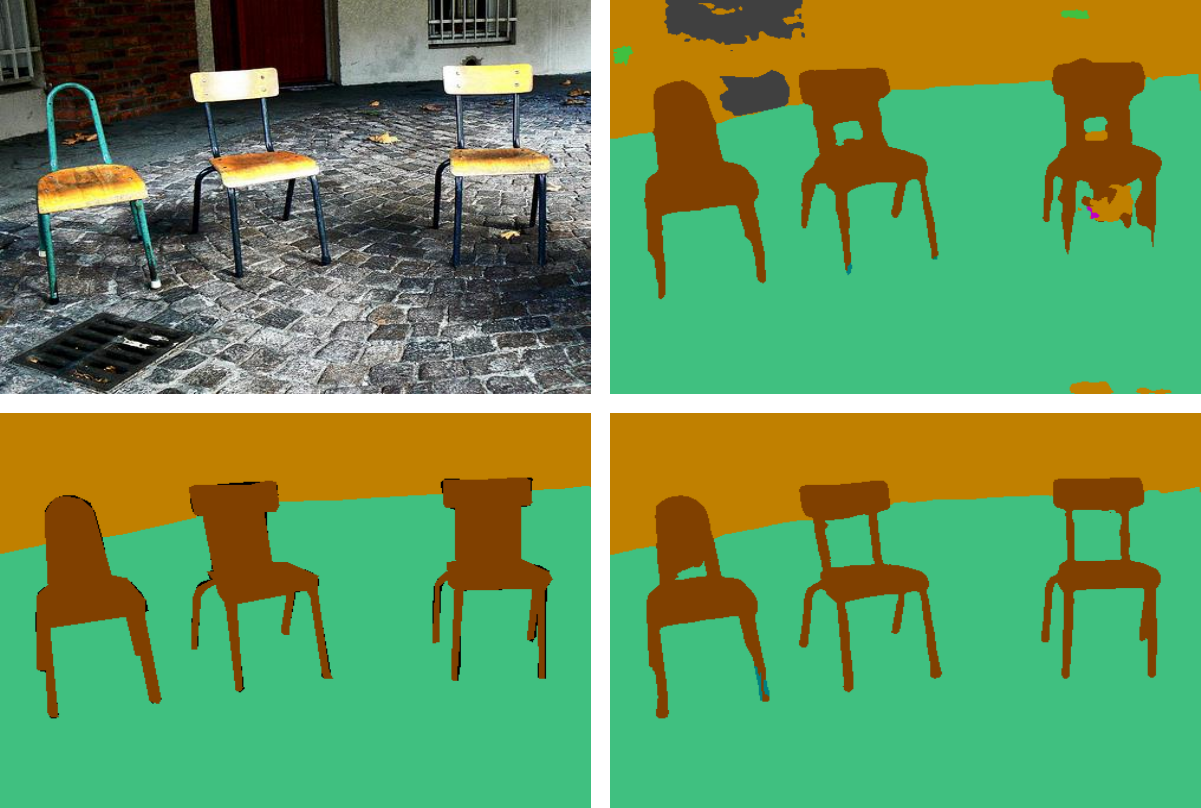}}
		\hspace{-3mm}
		\subfigure[] {
			\label{Fig.5.sub.4}
			\includegraphics[width=0.559\textwidth]{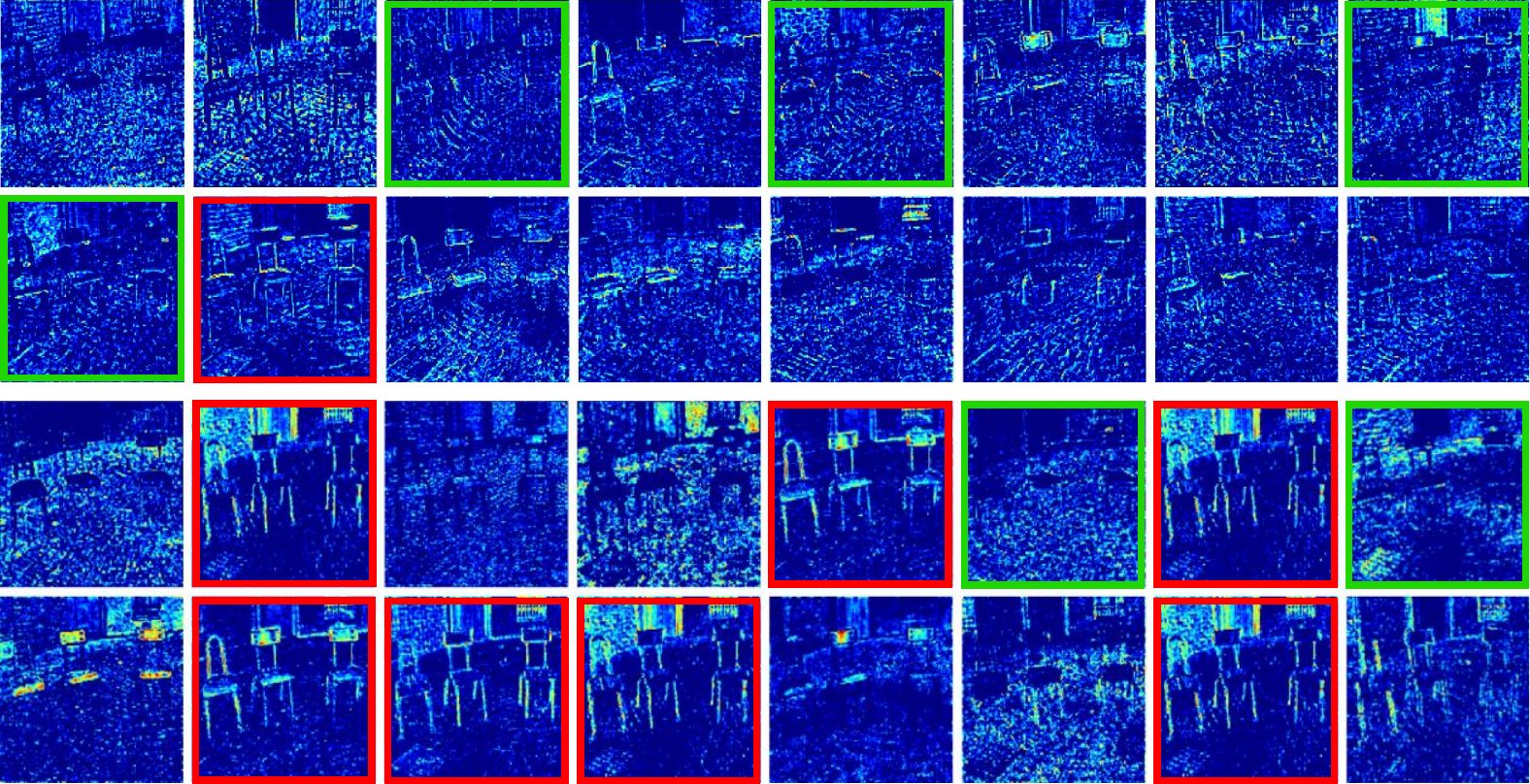}}
		
		\caption{Comparison of the segmentation results and the encoder features between FPN and AGLN. (a) and (c) show input images and segmentation maps (left-top: input image, left-bottom: groud truth, right-top: FPN, right-bottom: Ours.). (b) and (d) show the comparison of the encoder features before and after Local Refinement (First two rows: original encoder features, last two rows: refined encoder features.). Red box: valuable local details, green box: noisy features (best viewed in color).}
		
		\label{Fig.5}
	\end{figure*}

	\begin{figure*}[!t]
		\includegraphics[width=\textwidth]{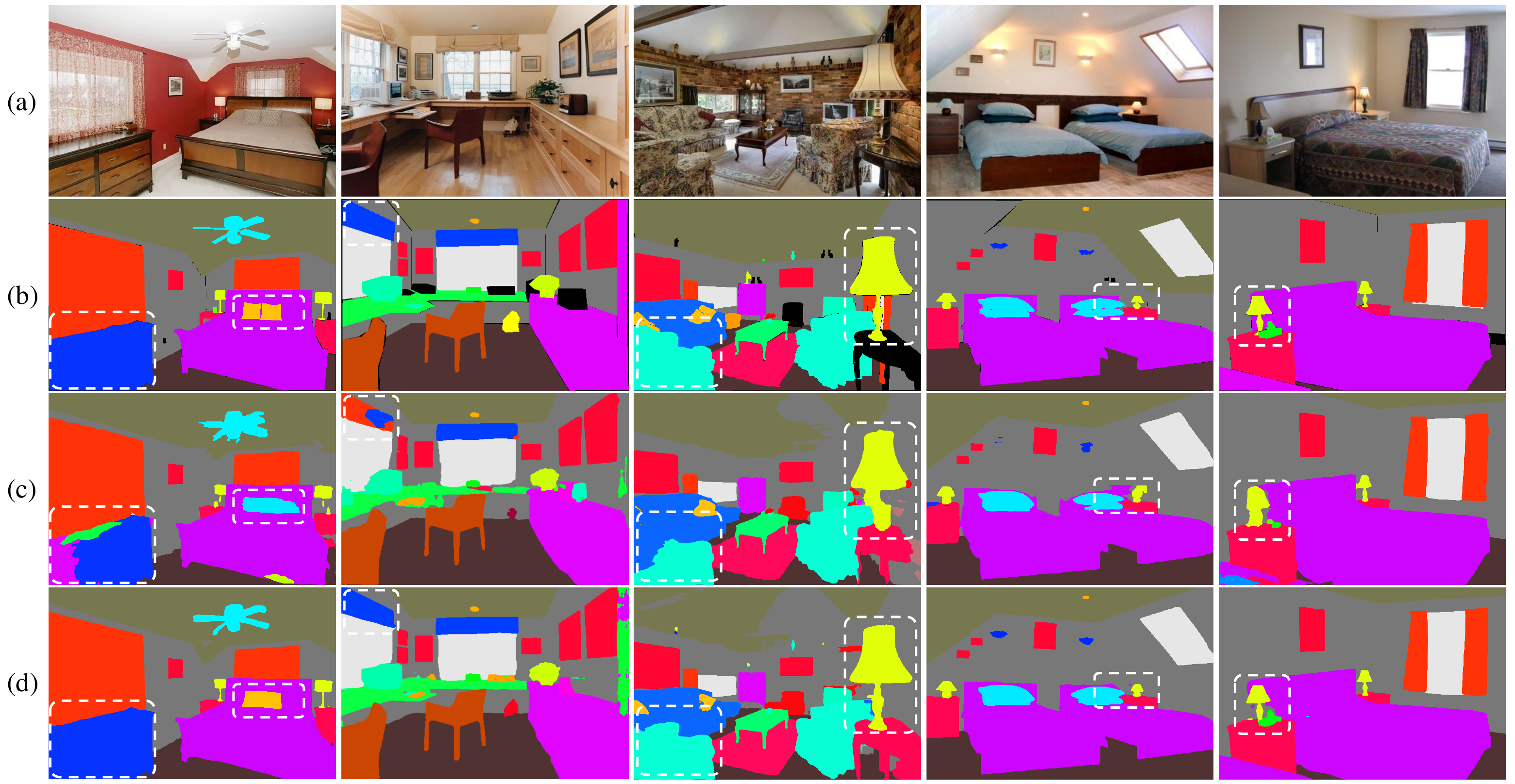}
		\caption{Example results on ADE20K validation set. The images in each column from top to bottom are (a) Input images, (b) Ground truth, (c) FPN, and (d) Ours. The main contrasts are marked by the white dashed lines (best viewed in color).}
		\label{Fig.6}
	\end{figure*}

	\begin{table}[!t]
		\renewcommand{\arraystretch}{1.3}
		\caption{The results of densely connected decoder and multi-scale evaluation. Dense represents the densely connected decoder and MS represents the multi-scale evaluation}
		\label{table4}
		\centering
		\begin{tabular}{ c c c c |c}
			\toprule[1.5pt]
			Method & Backbone & Dense & MS & mIoU$\%$ \\
			\hline
			FPN & ResNet-50 &  & & 46.91 \\
			\hline
			AGLN & ResNet-50 &  & & 51.98 (+5.07) \\
			AGLN & ResNet-50 & $\surd$ & & 52.45 (+5.54) \\
			AGLN & ResNet-50 & $\surd$ & $\surd$ & \textbf{53.42} (+6.51) \\
			\bottomrule[1.5pt]
		\end{tabular}
	\end{table}

	\begin{table*}[!t]
		\renewcommand{\arraystretch}{1.3}
		\caption{Performance (mIoU\%) and Complexity Comparison of the AGLN and AGLN-Lite. FCN-8s denotes that the output stride of the FCN encoder is 8}
		\label{table8}
		\centering
		\begin{tabular}{ c c | c c c | c c}
			\toprule[1.5pt]
			Method & Backbone & PASCAL Context & ADE20K & PASCAL VOC 2012 & Parameters & FLOPs \\
			\hline
			FCN-8s & D-ResNet-101 & 51.38 & 41.40 & 73.57 & 52.09 M & 196.75 G \\
			FPN & ResNet-101 & 52.82 & 42.58 & 81.26 & 54.32 M &  124.59 G \\
			AGLN & ResNet-101 & 56.23 & 45.38 & 84.89 & 61.71 M & 187.16 G \\
			AGLN-Lite & ResNet-101 & 56.21 & 44.04 & 84.38 & 55.74 M & 135.56 G \\
			\bottomrule[1.5pt]
		\end{tabular}
	\end{table*}

	\subsection{AGLN-Lite}
	Although the AGLN gains an impressive improvement in the segmentation performance compared to the baseline, it results in extra parameters and computational cost due to the additional $3\times3$ convolutions in the repeated Context Fusion Blocks, especially in the densely connected decoder. Therefore, a lightweight version of AGLN is developed by applying the $3\times3$ \textit{depth-wise separable convolutions} as the replacement of the original $3\times3$ convolutions in Context Fusion Blocks, namely AGLN-Lite. The comparison of our AGLN, AGLN-Lite, and FPN shown in Table \ref{table8} is performed with the same experimental setting, while the results of FCN are from the model zoo (best implementation) of MMSegmentation\footnote{https://github.com/open-mmlab/mmsegmentation/tree/master/conFig/fcn}. As can be seen, the AGLN-Lite gains a considerable reduction in parameters and FLOPs at the cost of a slight decrease in segmentation performance. Notably, taking Dilated-ResNet as the backbone, FCN-8s suffers from high computational consumption and so do its variants (i.e. the Dilated-FCN-based models). In comparison, based on the vanilla ResNet, our AGLN-Lite becomes a much better choice when considering the balance between segmentation accuracy and computation efficiency.
	
	\subsection{Understanding AGLN}
	\label{understanding}
	To illustrate the ability of AGLN, we visualize the enhanced decoder features produced by Global Enhancement Method and the refined encoder features predicted by Local Refinement Module.
	
	\subsubsection{Global Enhancement Visualization}
	Fig. \ref{Fig.4} depicts the semantic descriptor maps and the enhanced decoder features generated by  Global Enhancement Method on PASCAL Context.
	As shown in Fig. \ref{Fig.4.sub.3}, original feature maps from the last decoder layer of FPN show vague semantic boundaries and weak discriminations between different categories, resulting in misclassifications of large-scale objects (e.g., FPN results in Fig. \ref{Fig.4.sub.5}). Obviously, the semantic descriptor maps in Fig. \ref{Fig.4.sub.2} represent minor intra-class diversity within a certain object and large inter-class diversity between different targets. The main reason is that  the semantic descriptors are obtained by the global attention pooling, which owns a clustering effect and can aggregate the similar semantic concepts together. Moreover, each location of the decoder feature adaptively integrates the semantic descriptors based on their specific attention weights, distributing similar semantic concepts in a consistent semantic region. In Fig. \ref{Fig.4.sub.4}, the combinations of original decoder features and semantic descriptor maps (i.e. the enhanced decoder features) show much stronger spatial semantic consistency. The comparison of the segmentation results shown in Fig. \ref{Fig.4.sub.5}, \ref{Fig.4.sub.6}, and \ref{Fig.4.sub.7} demonstrates that the Global Enhancement Method helps to generate more semantically consistent and accurate segmentation results for large-scale objects. 
	
	\subsubsection{Local Refinement Visualization}
	Fig. \ref{Fig.5.sub.2} and \ref{Fig.5.sub.4} present the visualization samples from the original encoder feature maps (first two rows) and our refined encoder feature maps (last two rows). These visualized channels are randomly sampled from the highest-resolution (1/4 of the input size) encoder feature maps. In these figures, the red box denotes the valuable local details which may contribute to the segmentation of delicate object edges (e.g., the aircraft propeller in Fig. \ref{Fig.5.sub.1} and the chair legs in Fig. \ref{Fig.5.sub.3}). In contrast, the green box marks out the noisy features showing no visible distinction between foreground objects and background. 
	
	As shown in Fig. \ref{Fig.5.sub.2} and \ref{Fig.5.sub.4}, the effect of the refinement module is obvious. Specifically, the noisy features are filtered out and the fine-grained details are greatly enhanced. This remarkable improvement benefits from three aspects: First, the channel attention significantly suppresses the noisy channels of the encoder features for the gap between them and the semantically rich decoder features. Second, the informative channels of encoder features are greatly enhanced for their relatively high similarity with the enhanced decoder features. Third, the spatial gating map is employed to selectively enhance spatial details in a specific semantic area and suppress other irrelevant regions. The comparison of the segmentation results between the proposed AGLN and FPN (Fig. \ref{Fig.5.sub.1} and \ref{Fig.5.sub.3}) demonstrates the effectiveness of our Local Refinement Module.

	\section{Conclusion}
	In this paper, a novel AGLN is proposed to improve the encoder-decoder network for image segmentation. First, a Global Enhancement Method is designed to capture global semantic information from high-level features to complement the deficiency of global contexts in the upsampling process. Then, a Local Refinement Module is built to refine the noisy encoder features in both channel and spatial dimensions before the context fusion. After that, the proposed two methods are integrated into the Context Fusion Blocks, enabling the AGLN to generate semantically consistent segmentation masks on large-scale stuff and accurate boundaries on delicate objects. Furthermore, a lightweight implementation AGLN-Lite is developed, achieving a balance of performance and computational complexity. Experiments on the PASCAL Context, ADE20K, and PASCAL VOC 2012 datasets show that the proposed AGLN provides an effective solution to improve the performance of the FPN baseline and achieves state-of-the-art segmentation results ($56.23\%$ mean IOU on the PASCAL Context dataset). In conclusion, the proposed AGLN is simple, effective, and model-agnostic in the sense that it can be applied to various encoder-decoder architectures. 
	
	\section*{Acknowledgment}
	This work was supported by the Fundamental Research Funds for the China Central Universities of USTB (FRF-DF-19-002), Beijing Key Discipline Development Program (No. XK100080537).
	
	\ifCLASSOPTIONcaptionsoff
	\newpage
	\fi

	\bibliographystyle{IEEEtran}
	\bibliography{reference}
	
	%

\end{document}